%% file: acl_latex.tex
\documentclass[11pt]{article}

\usepackage[final]{acl}

\usepackage{times}
\usepackage{latexsym}
\usepackage[T1]{fontenc}
\usepackage[utf8]{inputenc}
\usepackage{microtype}
\usepackage{inconsolata}
\usepackage{natbib}
\usepackage{graphicx}
\usepackage{algorithm}
\usepackage{algorithmic}
\usepackage{amsmath,amsfonts}
\usepackage{array}
\usepackage[caption=false,font=normalsize,labelfont=sf,textfont=sf]{subfig}
\usepackage{textcomp}
\usepackage{verbatim}
\usepackage{microtype}
\usepackage{booktabs}
\usepackage{multirow}
\usepackage{multicol}
\usepackage{makecell}
\usepackage{enumerate}
\usepackage{xcolor}
\usepackage{xspace}
\usepackage{enumitem}
\usepackage{balance}
\usepackage{amsthm}
\usepackage{hyperref}
\usepackage{subcaption}
\usepackage{graphicx}
\usepackage{tabularx}
\usepackage{fontawesome5}
\usepackage[most]{tcolorbox}
\definecolor{c_context_bg}{RGB}{245, 247, 250} 
\definecolor{c_context_frame}{RGB}{66, 133, 244} 
\definecolor{c_fail_bg}{RGB}{255, 242, 242}     
\definecolor{c_fail_frame}{RGB}{219, 68, 55}    
\definecolor{c_ours_bg}{RGB}{236, 249, 240}     
\definecolor{c_ours_frame}{RGB}{15, 157, 88}    
\newtcolorbox{aibox}[2][]{
  colback=gray!5!white, 
  colframe=gray!75!black, 
  fonttitle=\bfseries,
  title={#2},
  #1
}
\newcommand{\model}{\texttt{Re$^{2}$A}\xspace}

\title{\texttt{Re$^{2}$A}: Situated Conversational Recommendation via Rubric-based Preference Reasoning and Alignment}

\author{
    Dongding Lin$^{1}$, ~~Jian Wang$^{2 \dagger}$, ~~Xiaoyan Zhao$^{3}$, ~~Wenjie Li$^{1}$ \\
    $^{1}$ Department of Computing, The Hong Kong Polytechnic University \\ 
    $^{2}$ College of Computer Science, Sichuan University \\
    $^{3}$ The Chinese University of Hong Kong \\
    \texttt{dongding88.lin@connect.polyu.hk} ~~~
    \texttt{wangjian51@scu.edu.cn} \\
    \texttt{xzhao@se.cuhk.edu.hk} ~~~ 
    \texttt{cswjli@comp.polyu.edu.hk}
}

\begin{document}
\maketitle

\renewcommand{\thefootnote}{$\dagger$}
\footnotetext[1]{Corresponding author.}
\setcounter{footnote}{0}
\renewcommand{\thefootnote}{\arabic{footnote}}

\begin{abstract}
\input{sections/0-Abstract}
\end{abstract}

\input{sections/1-Intro}
\input{sections/2-Task}
\input{sections/3-Method}
\input{sections/4-ExperimentalSetup}
\input{sections/5-ResultsAndAnalysis}
\input{sections/6-RelatedWork}
\input{sections/7-Conclusion}

\input{sections/limitations}

\bibliography{custom}

\newpage
\appendix
\input{sections/appendix}

\end{document}

%% file: sections/0-Abstract.tex
Real-world recommendation scenarios are commonly grounded in shared physical environments during user–recommender interactions.
This motivates situated conversational recommendation (SCR), a complex task requiring recommender assistants to jointly reason over dialogue history, co-observed scenes, and in-scene item attributes.
However, current approaches struggle with this setting due to two intertwined challenges: accurately understanding situated user preferences throughout the conversation and generating responses that simultaneously satisfy user needs and grounded situations. 
To this end, we propose \model, a framework that formulates SCR as a structured \textit{reason-then-align} process. 
We introduce rubric-based preference reasoning, which uses automated rubrics to guide the model toward producing explicit preference states. 
Based on these states, we propose a preference-conditioned optimization to align response generation with dual objectives: user preference satisfaction and situation consistency.
Extensive experiments on two SCR datasets demonstrate that \model consistently outperforms state-of-the-art methods, delivering more precise, context-aware conversational recommendations.
Our code is available at \url{https://github.com/DongdingLin/Re2A}.

%% file: sections/1-Intro.tex
\section{Introduction}
Conversational recommendation~\cite{li2018towards}, which aims to deliver personalized recommendations through natural language interactions, has been a pivotal research area in recent years. 
Most existing conversational recommendation systems primarily operate in a language-only setting~\cite{zhou2022c2,10.1145/3640457.3688146,DBLP:conf/naacl/YoonHEM24}, ignoring the physical environment in which users are located.
However, real-world recommendation scenarios, such as navigating live promotions in clothing or furniture stores, are inherently \textit{situated}, where effective communication relies on a shared observation of the surrounding environment~\cite{DBLP:conf/coling/MoonKCDPLWDBCSG20,DBLP:conf/emnlp/KotturMGD21}. 
This motivates the study of Situated Conversational Recommendation (SCR)~\cite{DBLP:conf/mm/LinWLL24,DBLP:journals/corr/abs-2412-18416,lin-etal-2026-reasoning}.
SCR focuses on multimodal interactions between users and conversational agents, which are required to recommend items by considering dialogue history, co-observed visual scenes, and in-scene item attributes. 
Shifting from a ``text-only'' to a ``scene-aware'' formulation, SCR transforms conversational agents from isolated recommenders into situated assistants~\cite{DBLP:conf/ecir/MukandeACDO24,zhang2025bootstrapping,vanderhoeven-etal-2025-trace}.

\begin{figure}[t!]
\centering
\includegraphics[width=1.0\linewidth]{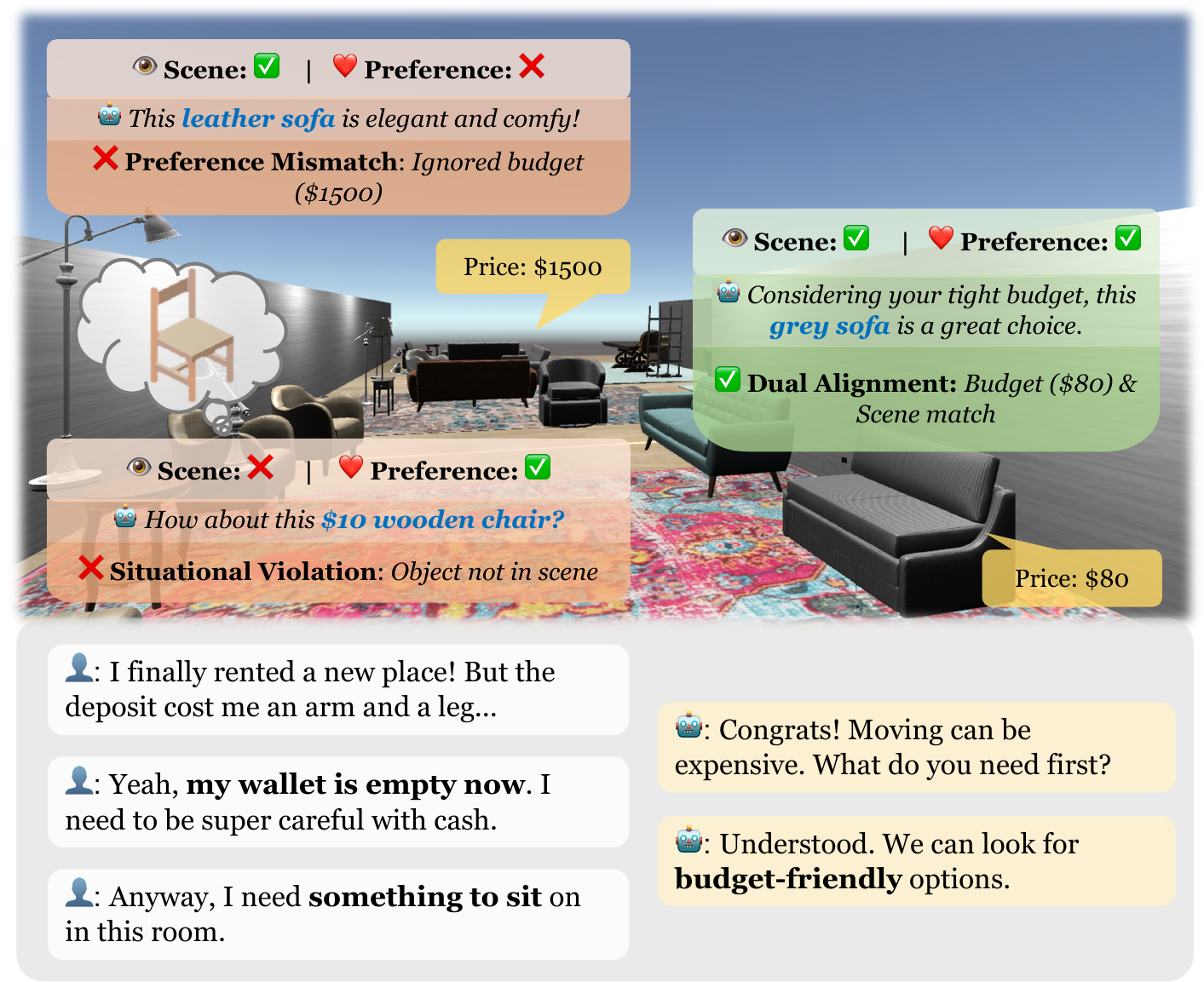}
    \caption{An illustrative example of situated conversational recommendation, where user preference reasoning and situation alignment are two key aspects.}
\label{fig:intro_example}
\vspace{-9pt}
\end{figure}

Despite its potential, SCR poses two fundamental challenges that distinguish it from conventional conversational recommendation, yet this direction remains underexplored.
As illustrated in Figure~\ref{fig:intro_example}, the first challenge lies in \textit{inferring users' situated preferences (or needs)}. 
Unlike text-only settings, user preferences in SCR are tightly coupled with the located scene and may depend on implicit needs, visually similar items, or even shift dynamically with the environment. 
The complexity of such multimodal contexts makes it difficult for models to accurately infer the user's underlying intent.
The second challenge concerns how to effectively \textit{leverage situated preferences and scene information to generate appropriate responses}. 
In SCR, recommendation responses are subject to dual objectives, as they must satisfy user preferences while grounding in visible candidate items.  
Addressing this challenge demands dual alignment with both the user and the environment, enabling agents to deliver recommendations that are not only relevant but also contextually appropriate.

In this paper, we propose \textbf{R}ubric-based Pr\textbf{e}ference \textbf{Re}asoning and \textbf{A}lignment (\textbf{\model}), a framework that formulates SCR as a \textit{reason-then-align} process.
Rather than introducing a new optimization algorithm, our core insight is to derive a structured preference state that serves as a shared interface among situated reasoning and response generation.
\model performs rubric-based preference reasoning, explicitly inferring intermediate user preference states from the dialogue history and co-observed scenes.
Inspired by the Chain-of-Thought reasoning~\cite{10.5555/3600270.3602070}, these preference states are represented in a structured textual form, making the reasoning process inspectable.
Unlike free-form CoT rationales, the preference state explicitly records the inferred user needs, attribute constraints, and visual target before any item is recommended.
As collecting verifiable supervision for such reasoning is costly, we employ context-aware, multi-dimensional rubrics to obtain rewards and optimize the model via policy optimization~\cite{shao2024deepseekmath}, encouraging robust preference reasoning under complex situational contexts.

Building upon the reasoned preference states, \model addresses the second challenge through dual-alignment response generation. 
We formulate response generation as a contrastive learning problem~\cite{rafailov2023direct} and introduce a user preference-conditioned optimization objective. 
By constructing heterogeneous negative responses that specifically violate either user preferences or scene constraints, the model learns to align its outputs with both situation consistency and user intent satisfaction. 
As a result, \model produces recommendations that are contextually appropriate.

Our contributions are summarized as follows:
\begin{itemize}[leftmargin=*]
    \item We propose \model, a novel framework for SCR that reformulates the task as a structured \textit{reason-then-align} process, explicitly separating situated preference reasoning from response alignment. The framework centers on a structured preference state that acts as a shared interface among preference reasoning, scene-constrained item recommendation, and response generation.
    \item We leverage rubric-guided policy optimization to derive explicit user preference states, while introducing user preference-conditioned optimization with rubric-based candidates to achieve dual alignment with both user intent and situational constraints. This avoids manual annotation of reasoning traces by leveraging automated rubric induction and synthetic preference construction.
    \item Experiments across two SCR datasets demonstrate that \model consistently outperforms competitive approaches, producing recommendations that better capture users' situated needs while remaining factually aligned with situations.
\end{itemize}

%% file: sections/2-Task.tex
\section{Task Definition}
\label{sec:task_definition}

We formulate SCR as a sequential decision-making process within a shared multimodal environment. The environment comprises a collection of visual scenes $\{\mathcal{S}_i\}_{i=1}^{M}$, where each scene $\mathcal{S}_i$ contains its own candidate item set $\mathcal{I}_i = \{o_{i,j}\}_{j=1}^{N_i}$. Crucially, each item $o_{i,j}$ is a multimodal entity, characterized jointly by its visual appearance (e.g., color), spatial coordinates, and structured metadata (e.g., price).
At the $t$-th turn, the agent receives a situated context $X_t = (\mathcal{S}_t, \mathcal{I}_t, \mathcal{C}_t)$, composed of the co-observed scene $\mathcal{S}_t$, scene-specific item set $\mathcal{I}_t$, and the dialogue context $\mathcal{C}_t = \{u_1, Y_1, \ldots, u_{t-1}, Y_{t-1}, u_t\}$, which includes past conversations and the current user utterance $u_t$.

The objective of SCR is to learn a mapping function that utilizes this context to simultaneously:
(1) identify the appropriate target item $o_t^* \in \mathcal{I}_t$ that satisfies the user's needs; and
(2) generate a natural language response $Y_t$ that is aligned with the user's preference while remaining grounded in the scene.

%% file: sections/3-Method.tex
\section{Method}

\begin{figure*}[t!]
\centering
\includegraphics[width=1.0\textwidth]{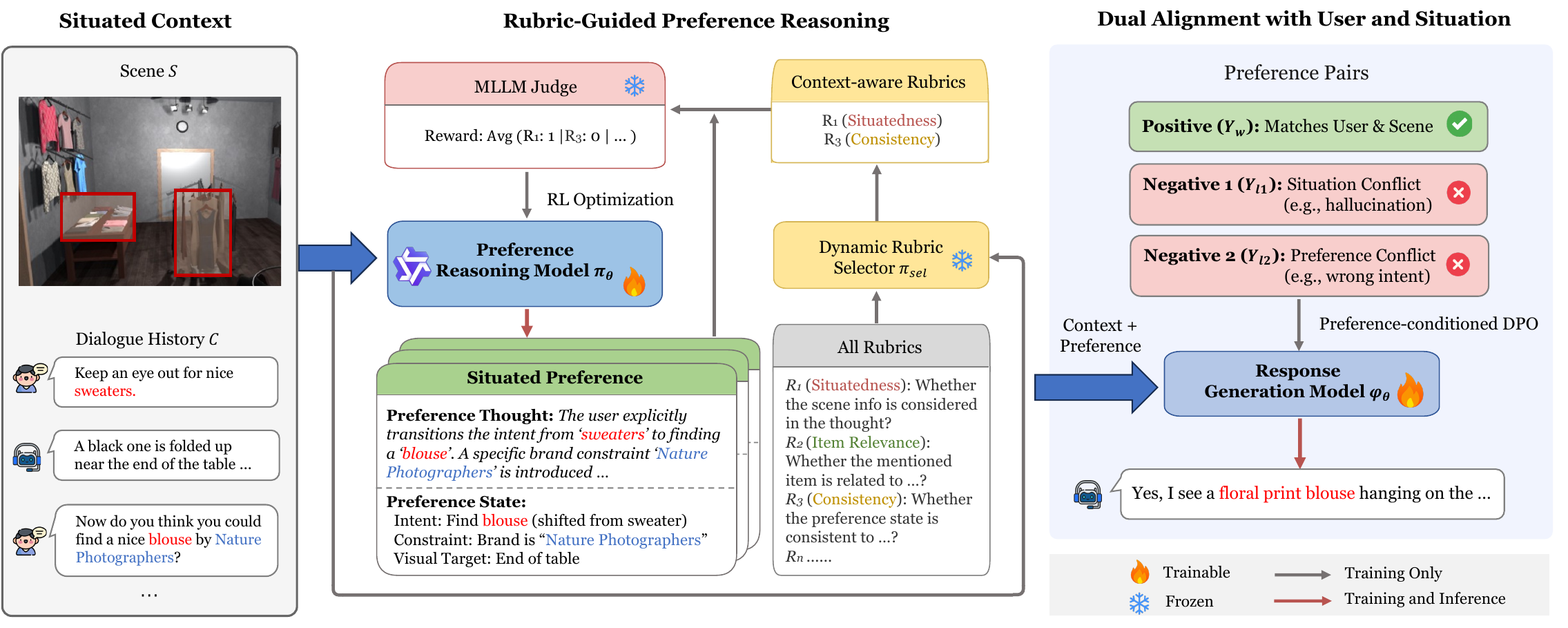}
    \caption{Overview of the proposed \model framework for situated conversational recommendation.}
\label{fig:framework}
\vspace{-8pt}
\end{figure*}

In this section, we present \textbf{R}ubric-based Pr\textbf{e}ference \textbf{Re}asoning and \textbf{A}lignment (\textbf{\model}) for situated conversational recommendation. Figure~\ref{fig:framework} shows the overview of \model, which follows a three-step pipeline: rubric-guided preference reasoning (see \S\ref{sec:preference_reasoning}), dual alignment with user and situation (see \S\ref{sec:dual_alignment}), and cascaded inference (see \S\ref{sec:inference}).

\subsection{Rubric-Guided Preference Reasoning}
\label{sec:preference_reasoning}
Accurately inferring latent user needs presents a fundamental challenge in SCR.
Relying solely on item labels to supervise preference reasoning risks capturing superficial correlations while failing to grasp the rationale determining how specific spatial or visual cues motivated the choice.
To bridge this gap, we pivot the reasoning objective from predicting the \textit{what} to explaining the \textit{why}.
Given that this intermediate reasoning lacks natural supervision, we introduce a rubric-based framework to provide verifiable and process-oriented guidance.
This design differs from generic CoT and conventional state tracking in two respects.
First, the agent leverages its thinking ability to yield a structured preference state that records the possible user needs, attribute constraints, and visual targets.
Second, since the intermediate state has no direct annotation, the turn-relevant rubrics evaluate each thinking process and its resulting state against evidence in the co-observed scene, rather than supervising individual reasoning steps.

\paragraph{Automated Rubric Induction.}
To derive scalable supervision without manually annotating preference-reasoning traces, we construct a global rubric set $\mathcal{R}$ via an induction process driven by a Multimodal Large Language Model (MLLM). 
Given a small curated seed dataset  $\mathcal{D}_{\text{seed}} = \{(\mathcal{S}_k, \mathcal{C}_k)\}_{k=1}^K$, and a constitutional meta-instruction $\mathcal{I}_{\text{meta}}$ (see Appendix~\ref{appendix:constitutional_meta_instruction}), we prompt the MLLM to yield reasoning criteria, formulating a rubric set. The rubric set is structured into the following four facets:  
\begin{equation}
\label{eq:rubric_set}
\mathcal{R} = \mathcal{R}_{\text{sit}} \cup \mathcal{R}_{\text{rel}} \cup \mathcal{R}_{\text{con}} \cup \mathcal{R}_{\text{aux}},
\end{equation}
where $\mathcal{R}_{\text{sit}}$ (\textit{Situatedness}) evaluates visual grounding (e.g., spatial references), $\mathcal{R}_{\text{rel}}$ (\textit{Item Relevance}) assesses alignment with the constraint of item attributes, $\mathcal{R}_{\text{con}}$ (\textit{Consistency}) ensures logical coherence with dialogue history, and $\mathcal{R}_{\text{aux}}$ (\textit{Auxiliary}) captures domain-specific criteria beyond the prior three facets identified by the MLLM.
Crucially, each rubric $r \in \mathcal{R}$ is formulated as a binary predicate function $r(\cdot) \to \{0, 1\}$, enabling standardized binary evaluation by a frozen MLLM judge (see Appendix~\ref{appendix:rubrics_examples} for concrete examples).
These four facets follow SCR's core requirements of visual grounding, constraint satisfaction, and multi-turn coherence; a manual coverage check on 50 sampled dialogue turns finds that all turn-level requirements map onto these facets (Appendix~\ref{appendix:audits}).

\paragraph{Dynamic Rubric Selection.}
Uniformly applying the global rubric set $\mathcal{R}$ across dynamic dialogue turns introduces noisy supervision, as irrelevant constraints obscure the reasoning focus. 
To this end, we employ a dynamic rubric selector, instantiated by a frozen MLLM ($\pi_{\text{sel}}$),  to select a pertinent subset $\mathcal{R}_{\text{act}} \subset \mathcal{R}$ conditioned on the situated context $X_t$. By analyzing the primary communicative needs, $\pi_{\text{sel}}$ gates inactive criteria. For instance, a purely spatial query such as ``\textit{the one on the left}'' primarily activates situatedness rubrics, whereas a mixed query such as ``\textit{the red one on the left}'' activates both situatedness and item-relevance rubrics. 
This selective activation provides denser and less noisy supervision. The prompting template is provided in Appendix~\ref{appendix:selector_prompt}.

\paragraph{Preference State Generation with Rubric-based Rewards.}
To bridge raw multimodal context and structured preference inference, we train a preference reasoning model $\pi_{\theta}$ to generate user preference reasoning traces explicitly. 
Given the situated context $X$ and a structural instruction $\mathcal{I}_{\text{reason}}$ (see Appendix~\ref{appendix:reasoning_prompt}), $\pi_{\theta}$ produces a \emph{preference thought} $\mathcal{T}$, which captures intermediate reasoning over preference shifts and spatial constraints, followed by a structured \emph{preference state} $\mathcal{P}$ that formalizes the inferred user needs, attribute constraints, and visual targets into a structured textual format for prompt-based item recommendation and rubric verification.

Since such preference states are latent and lack direct supervision, we adopt a group-based learning strategy to provide relative training signals. During each training step, $\pi_{\theta}$ samples a group of $G$ reasoning candidates $\{\mathcal{O}_1, \ldots, \mathcal{O}_G\}$, where each $\mathcal{O}_g = (\mathcal{T}_g, \mathcal{P}_g)$. 
At $t$-th turn, the dynamically selected rubric subset $\mathcal{R}_{\text{act}}$ serves as the evaluation standard. We use MLLM-as-a-Judge to assess whether each candidate satisfies the active rubrics, yielding a set of binary verification outcomes.
Formally, we instantiate this judge as a binary evaluator $f_{\text{judge}}:(\mathcal{O}_g,r,X_t)\rightarrow\{0,1\}$ using a frozen Qwen3-VL-8B-Instruct~\citep{bai2025qwen3vltechnicalreport} model.
For each active rubric, the evaluator receives the situated context, candidate preference thought, preference state, rubric question, and its evidence indicators, then it outputs a structured JSON verdict with supporting evidence; a deterministic parser maps \texttt{YES} to 1 and \texttt{NO} to 0.
The exact judging prompt is provided in Appendix~\ref{appendix:judge_prompt}.

Concretely, each rubric $r \in \mathcal{R}_{\text{act}}$ functions as a binary criterion. The scalar reward $R_g$ for candidate $\mathcal{O}_g$ is computed as the satisfaction rate:
\begin{equation}
    \label{eq:rubric_reward}
    R_g = \frac{1}{|\mathcal{R}_{\text{act}}|} \sum_{r \in \mathcal{R}_{\text{act}}} 
    f_{\text{judge}}(\mathcal{O}_g, r, X_t).
\end{equation}
Rather than relying on absolute reward values, we compute the relative advantage to reduce variance:
\begin{equation}
    A_g = \frac{R_g - \mu_R}{\sigma_R + \epsilon},
\end{equation}
where $\mu_R$ and $\sigma_R$ denote the mean and standard deviation of rewards $\{R_1, \ldots, R_G\}$ within the group, and $\epsilon$ ensures numerical stability.

We optimize $\pi_{\theta}$ using Group Relative Policy Optimization (GRPO)~\cite{shao2024deepseekmath}. By leveraging group-normalized advantages, GRPO not only stabilizes optimization to favor reasoning traces that satisfy active situated constraints, but also eliminates the need for an explicit value network, thereby significantly reducing the memory overhead of training on high-dimensional multimodal inputs. The objective function is defined as:
\begin{equation}
\begin{aligned}
\mathcal{J}&(\theta) = \frac{1}{G} \sum_{g=1}^G \bigg( \min \bigg( \rho_g(\theta) A_g, \text{clip}  (\rho_g(\theta), \\ & 1-\varepsilon, 1+\varepsilon) A_g \bigg) \ - \beta\cdot\mathbb{D}_{\text{KL}}(\pi_{\theta} || \pi_{\text{ref}}) \bigg), \end{aligned}
\end{equation}
where $\rho_g(\theta) = \frac{\pi_{\theta}(\mathcal{O}_g \mid X)}{\pi_{\text{old}}(\mathcal{O}_g \mid X)}$ is the probability ratio, $\pi_{\text{ref}}$ is the reference policy (initialized from the supervised baseline), and $\beta$ controls the KL-divergence penalty to prevent policy collapse.

\subsection{Dual Alignment with User and Situation}
\label{sec:dual_alignment}
Building upon the preference state $\mathcal{P}$ inferred by $\pi_{\theta}$, we aim to generate responses simultaneously aligned with user preferences and grounded in the scene.
We note that responses satisfying user preferences risk violating physical constraints (e.g., hallucinating absent items), whereas strictly scene-consistent responses may fail to capture the user’s actual needs.
Therefore, we formulate response generation as a conditional alignment problem.
Specifically, we propose a \emph{user preference-conditioned DPO} method for response generation, building on top of vanilla DPO~\citep{rafailov2023direct}, where the optimization trajectory is explicitly guided by the inferred preference state.

\paragraph{Heterogeneous Preference Pair Construction.}
To provide fine-grained supervision, we construct heterogeneous negative responses targeting distinct failure modes in SCR. For each ground-truth response $Y_w$, we leverage an LLM to automatically synthesize two types of negative samples using specific instruction templates (see Appendix~\ref{app:neg_prompts}):
\textbf{(i) Situation-conflicting negatives ($Y_{l1}$)} are constructed by injecting hallucinations that contradict the observed scene $\mathcal{S}$, such as recommending absent items or referencing non-existent spatial landmarks. These pairs explicitly penalize violations of visual grounding.
\textbf{(ii) Preference-conflicting negatives ($Y_{l2}$)} are generated by substituting the target item in $Y_w$ with a distractor item that is present in the scene but contradicts the inferred preference state $\mathcal{P}$ (e.g., violating price or attribute constraints). These pairs penalize responses that are grounded yet misaligned with user needs.
A post-hoc validation with both model and human evaluators confirms that the synthesized negatives satisfy the conflicts (see Appendix~\ref{appendix:audits}).

\paragraph{Alignment Training.}
We adopt two-stage alignment training to optimize the response generation policy $\varphi_{\theta}$. 
We condition the response generator during training on the ground-truth target item $o^*$ as a teacher-forced grounding anchor. This use of $o^*$ supervises how the generator verbalizes a recommendation once the target item is specified, rather than supervising the preference-reasoning stage described in \S\ref{sec:preference_reasoning}. 
During inference, the oracle anchor is unavailable and is replaced by the retrieved item. 
Thus, the training-inference discrepancy is confined to the source of the grounding anchor, while the generator is consistently conditioned on an explicit item in both phases.

Concretely, we first perform supervised fine-tuning. Given the fully grounded context $\tilde{X} = (\mathcal{S}, \mathcal{C}, \mathcal{P}, o^*)$, we maximize the likelihood of the response $Y_w$:
\begin{equation}
\mathcal{L}_{\text{SFT}}(\theta)
= -\mathbb{E}_{(\tilde{X}, Y_w) \sim \mathcal{D}}
\big[\log \varphi_{\theta}(Y_w \mid \tilde{X})\big].
\end{equation}
This yields a reference policy $\varphi_{\text{ref}}$. Subsequently, we employ DPO using the constructed heterogeneous pairs. The objective is defined as:
\begin{equation}
\begin{aligned}
&\mathcal{L}_{\text{DPO}}(\theta)
= -\mathbb{E}_{(\tilde{X},Y_w,Y_l)\sim\mathcal{D}_{\text{mix}}}
\Big[
\log \sigma \Big(\\
&\beta_{\text{DPO}} \big(\log \frac{\varphi_{\theta}(Y_w\mid \tilde{X})}{\varphi_{\text{ref}}(Y_w\mid \tilde{X})} - \log \frac{\varphi_{\theta}(Y_l\mid \tilde{X})}{\varphi_{\text{ref}}(Y_l\mid \tilde{X})}\big)
\Big)
\Big],
\end{aligned}
\end{equation}
where $\beta_{\text{DPO}}$ controls the KL-divergence penalty relative to $\varphi_{\text{ref}}$.
Training on $\mathcal{D}_{\text{mix}}$, a mixture of situation-centric $(Y_w, Y_{l1})$ and preference-centric $(Y_w, Y_{l2})$ pairs, enforces dual constraints: 
$Y_{l1}$ penalizes referencing absent entities (grounding), while $Y_{l2}$ ensures consistency with the preference state and its corresponding target anchor over scene-valid distractors (alignment).
Consequently, $\varphi_{\theta}$ converges to a policy simultaneously grounded in the scene and aligned with the user preference.

\subsection{Cascaded Inference}
\label{sec:inference}
To translate reasoning-derived preferences into grounded recommendations at inference time, \model adopts a cascaded strategy that decouples item retrieval from response generation. In this design, the inferred preference state $\mathcal{P}$ serves as a pivot, transforming implicit user needs into explicit retrieval constraints before response generation.

\paragraph{Item Recommendation.}
We identify target items through a constrained preference-to-item parser. Given the generated preference state $\mathcal{P}$, the parser ranks visible candidates in $\mathcal{I}$ according to how well their visual attributes, spatial positions, and metadata satisfy the explicit constraints in $\mathcal{P}$. 
In practice, this parser is instantiated with the same backbone model under a deterministic ranking prompt (see Appendix~\ref{app:inference_prompts}); it does not introduce an additional learned ranking objective, but operationalizes the reasoned preference state by mapping it onto the scene-specific candidate item set. This keeps item selection more transparent and restricts the search space to physically visible items.

\paragraph{Aligned Response Generation.}
From the ranked candidates, we select the top-ranked item $\hat{o}$ to drive response generation. 
Conditioned on the situated context $(\mathcal{S}, \mathcal{C}, \mathcal{P}, \hat{o})$, the model $\varphi_{\theta}$ generates an appropriate response at each turn. 
This explicit conditioning anchors the narrative to the specific target $\hat{o}$, ensuring the response is simultaneously grounded in the surrounding scene and aligned with user preferences.

%% file: sections/4-ExperimentalSetup.tex
\section{Experimental Setup}

\begin{table*}[t!]
\centering
\resizebox{0.99\textwidth}{!}{
\begin{tabular}{cl cccc cccc}
\toprule
\multirow{2}{*}[-0.6ex]{\textbf{Backbone}} &
\multirow{2}{*}[-0.6ex]{\textbf{Method}} &
\multicolumn{4}{c}{\textbf{SIMMC 2.1}} & \multicolumn{4}{c}{\textbf{SCREEN}} \\
\cmidrule(lr){3-6} \cmidrule(lr){7-10}   
&  & \textbf{Hit@1} & \textbf{Recall@5} & \textbf{MRR@5} & \textbf{NDCG@5} &
\textbf{Hit@1} & \textbf{Recall@5} & \textbf{MRR@5} & \textbf{NDCG@5}\\
\cmidrule(lr){1-2} \cmidrule(lr){3-6} \cmidrule(lr){7-10}

\multirow{7}{*}{LLaVA-NeXT} &
\multicolumn{1}{l}{Vanilla Prompting} & 11.63 & 12.08 & 11.80 & 11.87 & 14.81 & 15.29 & 14.99 & 15.06 \\
& \multicolumn{1}{l}{ICL} & 14.02 & 15.71 & 14.65 & 14.91 & 16.38 & 18.06 & 17.00 & 17.27 \\
& \multicolumn{1}{l}{CoT} & 13.59 & 14.84 & 14.05 & 14.25 & 15.77 & 18.52 & 16.79 & 17.22 \\
& \multicolumn{1}{l}{SFT} & 23.14 & 30.67 & 25.93 & 27.11 & 24.96 & 30.13 & 26.87 & 27.68 \\
& \multicolumn{1}{l}{CRAG} & 27.91 & 38.04 & 31.86 & 33.43 & 29.48 & 39.15 & 34.05 & 35.38 \\
& \multicolumn{1}{l}{ReGeS} & 32.18 & 56.89 & 39.09 & 43.34 & 35.29 & 61.94 & 41.10 & 45.99 \\
\cmidrule(lr){2-2} \cmidrule(lr){3-6} \cmidrule(lr){7-10}
& \multicolumn{1}{l}{\textbf{\model (Ours)}} & \textbf{40.35} & \textbf{60.37} & \textbf{48.83} & \textbf{51.74} & \textbf{41.03} & \textbf{65.81} & \textbf{52.03} & \textbf{55.53} \\

\cmidrule(lr){1-2}\cmidrule(lr){3-6} \cmidrule(lr){7-10}

\multirow{7}{*}{Qwen3-VL} &
\multicolumn{1}{l}{Vanilla Prompting} & 15.37 & 17.91 & 16.52 & 16.88 & 19.62 & 21.03 & 20.14 & 20.36 \\
& \multicolumn{1}{l}{ICL} & 18.25 & 22.06 & 20.16 & 20.65 & 22.17 & 25.91 & 23.55 & 24.14 \\
& \multicolumn{1}{l}{CoT} & 17.55 & 19.53 & 18.42 & 18.70 & 21.41 & 25.76 & 23.02 & 23.70 \\
& \multicolumn{1}{l}{SFT} & 31.24 & 39.14 & 33.20 & 34.61 & 34.58 & 40.71 & 36.58 & 37.58 \\
& \multicolumn{1}{l}{CRAG} & 35.64 & 45.31 & 38.62 & 40.28 & 37.09 & 48.35 & 40.18 & 42.13 \\
& \multicolumn{1}{l}{ReGeS} & 39.45 & 58.74 & 43.78 & 47.30 & 41.72 & 64.19 & 50.03 & 53.56 \\
\cmidrule(lr){2-2} \cmidrule(lr){3-6} \cmidrule(lr){7-10}
& \multicolumn{1}{l}{\textbf{\model (Ours)}} & \textbf{48.12} & \textbf{65.08} & \textbf{54.22} & \textbf{56.89} & \textbf{49.63} & \textbf{70.47} & \textbf{58.44} & \textbf{61.52} \\

\bottomrule
\end{tabular}}
\caption{Evaluation results of different preference reasoning methods for situated conversational recommendation. The best results are highlighted in \textbf{bold}.}
\label{table:recommendation_result}
\vspace{-8pt}
\end{table*}

\paragraph{Datasets.}
\label{sec:datasets}
We evaluate \model on two typical SCR datasets: \textbf{SIMMC 2.1}~\citep{kottur2023overview} and \textbf{SCREEN}~\citep{DBLP:conf/mm/LinWLL24}.
SIMMC 2.1 features photorealistic 3D scenes in fashion and furniture domains, presenting significant challenges in visual grounding and coreference resolution within dense visual environments. In contrast, SCREEN focuses on implicit preference reasoning, containing synthesized dialogues with evolving personas and complex attribute constraints that require multi-turn logic tracking. 
We unify these sources into standard tuples for consistent training. Detailed statistics are provided in Appendix~\ref{app:dataset_stats}.
The average dialogue contains roughly 10 turns, so the textual history itself can typically fit within modern 8K $\sim$ 32K context windows.
The central difficulty instead comes from visual density, where each scene contains 19.7 objects on average, forcing the model to distinguish fine-grained spatial and attribute cues from many scene-valid distractors.

\paragraph{Backbone Models.} 
To validate our \model framework across varying model architectures, we implement \model on two state-of-the-art open-source MLLMs: \textbf{LLaVA-NeXT}~\citep{liu2024llavanext} and \textbf{Qwen3-VL}~\citep{bai2025qwen3vltechnicalreport}.
We select LLaVA-NeXT to exploit its dynamic high-resolution image encoding, a critical feature for discerning fine-grained visual attributes in dense SCR scenes (e.g., distinguishing similar furniture). 
Additionally, we employ Qwen3-VL to harness its superior instruction-following and logical reasoning capabilities, ensuring robust adherence to complex, multi-turn attribute and spatial constraints.

\paragraph{Baseline Methods.}
We compare \model against three distinct types of methods.
\textit{(1) Inference-Only Strategies:} We employ Vanilla Prompting, In-Context Learning (ICL), and Chain-of-Thought (CoT) to probe the intrinsic reasoning capabilities of the backbones without parameter updates. 
\textit{(2) Multimodal Fine-Tuning:} We establish standard SFT on the target datasets as the foundational baseline for learnable methods. 
\textit{(3) Adapted Conversational Recommendation Methods:} We adapt two cutting-edge conversational recommendation methods for the SCR setting: CRAG~\citep{zhu2025collaborative}, which utilizes collaborative retrieval to bridge semantic gaps, and ReGeS~\citep{yang2025reges}, which optimizes candidate selection through reciprocal retrieval-generation synergy. For a fair comparison, all learnable baselines are equipped with the same multimodal backbones.

\begin{table}[t!]
\centering
\resizebox{0.95\linewidth}{!}{
\setlength{\tabcolsep}{2pt}
\begin{tabular}{@{}llcccc@{}}
\toprule
\textbf{Data} & \textbf{Method} & \textbf{Judge-Avg} $\uparrow$ & \textbf{SR} $\uparrow$ & \textbf{VHR} $\downarrow$ & \textbf{Human-Avg} $\uparrow$ \\
\midrule
\multirow{7}{*}{SIMMC}
& Vanilla & 5.29 & 15.1 & 45.7 & 0.94 \\
& ICL & 6.02 & 17.7 & 39.6 & 1.11 \\
& CoT & 6.32 & 16.4 & 38.4 & 1.2 \\
& SFT & 7.82 & 29.6 & 25.4 & 1.47 \\
& CRAG & 8.17 & 34.3 & 19.7 & 1.57 \\
& ReGeS & 8.49 & 38.6 & 15.2 & 1.65 \\
& \textbf{\model} & \textbf{9.31} & \textbf{46.8} & \textbf{5.2} & \textbf{1.87} \\
\midrule
\multirow{7}{*}{SCREEN}
& Vanilla & 5.49 & 16.9 & 42.1 & 0.97 \\
& ICL & 6.16 & 19.7 & 36.4 & 1.15 \\
& CoT & 6.51 & 18.4 & 34.9 & 1.24 \\
& SFT & 7.94 & 31.7 & 24.1 & 1.51 \\
& CRAG & 8.3 & 36.1 & 18.6 & 1.61 \\
& ReGeS & 8.56 & 40.9 & 14.6 & 1.7 \\
& \textbf{\model} & \textbf{9.19} & \textbf{48.6} & \textbf{5.1} & \textbf{1.86} \\
\bottomrule
\end{tabular}}
\caption{Response generation performance with the Qwen3-VL backbone. Full evaluation results are provided in Appendix~\ref{appendix:full_response_quality}.}
\label{table:response_summary}
\vspace{-8pt}
\end{table}

\paragraph{Evaluation Protocols.}
\label{sec:protocols}
We employ a multidimensional protocol to evaluate recommendation accuracy and response generation quality.
For \textbf{recommendation}, we report Hit@1, Recall@5, MRR@5, and NDCG@5 to measure ranking alignment with ground-truth items.
All inference-time results are obtained with the full cascaded pipeline: the response generator is conditioned on the retrieved item $\hat{o}$ produced by preference-to-item parsing, rather than the oracle target $o^*$ used as a teacher-forced grounding anchor during response-generator training.
Since no manual correction is applied to predicted preference states or resolved items, errors propagated from upstream stages are already reflected in all reported results.

For \textbf{response generation}, we adopt a three-tiered strategy:
(1) \textbf{LLM-as-a-Judge (turn-level):} We employ GPT-5.2 to score generated responses (1--10) on \textit{visual fidelity} (VF) for hallucination detection, \textit{constraint compliance} (CC) for preference satisfaction, and \textit{conversational helpfulness} (CH).
(2) \textbf{User simulation (dialogue-level):} Through interaction with a simulator (max 20 turns), we report \textit{success rate} (SR). Crucially, to diagnose the alignment dilemma, we measure \textit{weak-match rate} (WMR) and \textit{visual hallucination rate} (VHR).
(3) \textbf{Human evaluation:} Annotators rate responses (0 $\sim$ 2) on \textit{situational grounding} (Ground.), \textit{preference alignment} (Align.), and \textit{logical coherence} (Coher.).
Appendix~\ref{appendix:metric_details} provides detailed definitions and prompts.
Appendix~\ref{appendix:evaluator_robustness} further verifies the robustness of these protocols with independent judges and alternative user simulators.

\paragraph{Implementation Details.}
We instantiate \model with LLaVA-NeXT~\citep{liu2024llavanext} and Qwen3-VL~\citep{bai2025qwen3vltechnicalreport}, training only LoRA adapters while freezing visual modules. GPT-5.2 supports offline rubric and negative construction, and a frozen Qwen3-VL serves as the training-time selector and judge. Full optimization, decoding, and infrastructure settings are provided in Appendix~\ref{appendix:implementation_details}.

%% file: sections/5-ResultsAndAnalysis.tex
\section{Results and Analyses}

\subsection{Main Results}
\label{sec:main_results}

\paragraph{Recommendation Performance.}
Table~\ref{table:recommendation_result} shows that \model achieves the best performance across metrics.
On SIMMC 2.1, it surpasses the strongest baseline ReGeS by 8.17 percentage points in Hit@1 (40.35\% vs. 32.18\%, LLaVA-NeXT).
We attribute these gains to two factors:
(1) \textit{Overcoming ``Reasoning Deficit'':} Standard CoT consistently underperforms ICL (e.g., 17.55\% vs. 18.25\% with Qwen3-VL) and often drifts into ``hallucinated reasoning'' by inventing visual details; rubric-guided verification anchors \model to the scene.
(2) \textit{Logical Reasoning vs. Retrieval:} Retrieval baselines (CRAG, ReGeS) struggle with spatial negation or conditional attributes because implicit semantic similarity misses logical constraints. The structured \textit{reason-then-align} process models these constraints explicitly, improving grounding.
Thus, long-context access alone is insufficient: despite short dialogue histories, dense scenes with visually plausible distractors still cause prompt-based methods to drift toward hallucinated reasoning.

We note that Table~\ref{table:recommendation_result} compares complete systems under their intended training objectives, and thus does not by itself isolate the contribution of our training objectives from the synthesized supervision used by \model.
As a call-matched diagnostic, prompting each frozen baseline to first generate an intermediate representation improves both CRAG and ReGeS, with the structured preference state outperforming free-form reasoning under the same call budget (Appendix~\ref{appendix:matched_supervision}), indicating that the structured interface itself contributes beyond synthesized training data.

\paragraph{Response Generation Quality.}
Beyond recommendation accuracy, \model generates more grounded responses. 
Table~\ref{table:response_summary} summarizes response-quality results; full SIMMC 2.1 and SCREEN breakdowns are provided in Appendix~\ref{appendix:full_response_quality}.
On SIMMC 2.1, \model improves visual fidelity over ReGeS under Qwen3-VL (see Appendix Table~\ref{tab:simmc_response_quality}).
These metrics use the full cascaded setting with retrieved anchors $\hat{o}$, so item-resolution errors remain reflected in generation.
Moreover, preference-conditioned DPO resolves the \textit{alignment dilemma}: SFT exhibits high VHR by inventing attributes to force satisfaction, whereas \model reduces VHR by 20.2\% (25.4\% $\to$ 5.2\%) and Weak-Match Rate by 15.2\% (22.6\% $\to$ 7.4\%). 
This confirms that \model aligns responses with user preference without compromising physical fidelity.

\begin{figure}[t!]
    \centering
    \begin{minipage}[b]{0.49\linewidth}
        \centering
        \includegraphics[width=\linewidth]{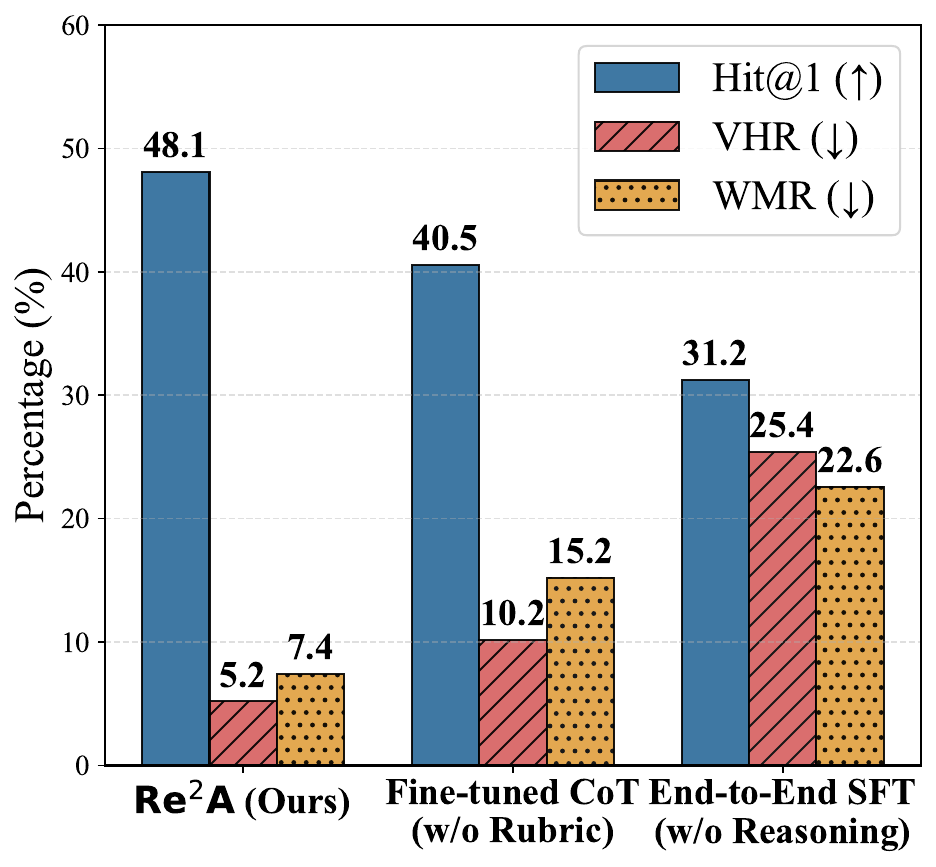}
        \centerline{\small \textbf{(a)} Reasoning Analysis}
    \end{minipage}
    \hfill
    \begin{minipage}[b]{0.49\linewidth}
        \centering
        \includegraphics[width=\linewidth]{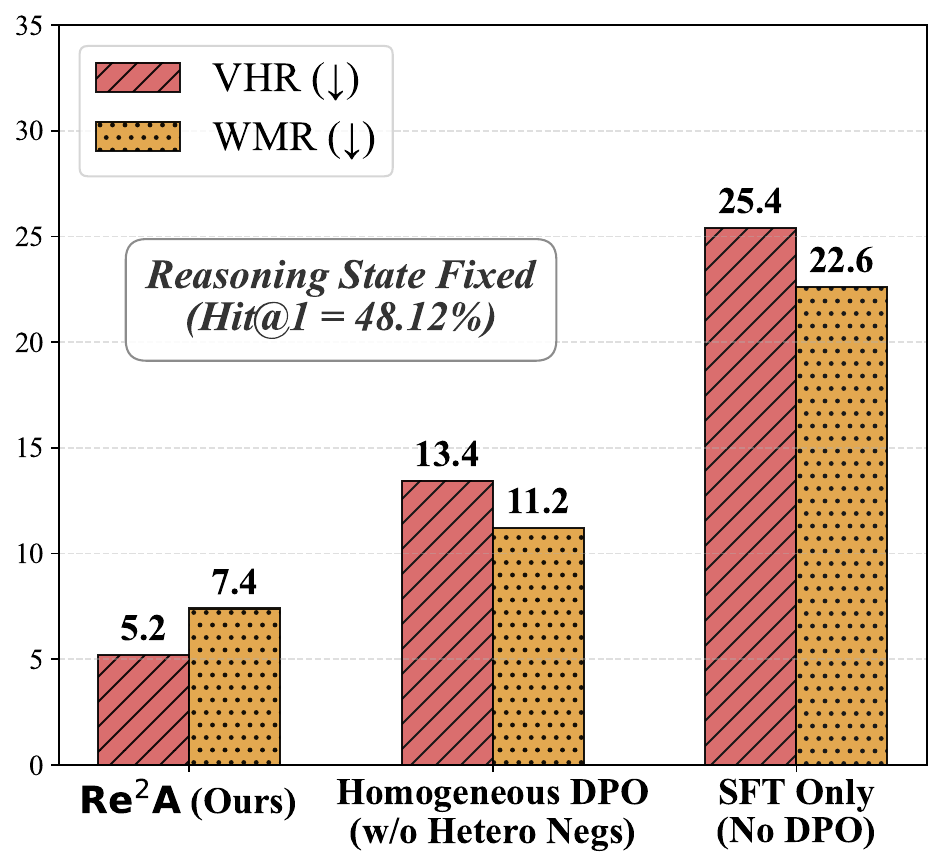}
        \centerline{\small \textbf{(b)} Alignment Analysis}
    \end{minipage}
    
    \vspace{-5pt}
    \caption{
        Ablation studies on SIMMC 2.1. 
        \textbf{(a)} Our rubric-guided reasoning yields higher accuracy (Hit@1) than baseline methods.
        \textbf{(b)} Heterogeneous DPO achieves the lowest hallucination rates.
    }
    \label{fig:ablation}
    \vspace{-6pt}
\end{figure}

\begin{table*}[t!]
\centering
\resizebox{0.95\textwidth}{!}{
\begin{tabular}{lccccccc}
\toprule
\textbf{Method} & \textbf{Preference Reasoning} & \textbf{Selector/Judge} & \textbf{Retrieval} & \textbf{Generation} & \textbf{Total} & \textbf{Hit@1} $\uparrow$ & \textbf{VHR} $\downarrow$ \\
& \textbf{(ms)} & \textbf{(ms)} & \textbf{(ms)} & \textbf{(ms)} & \textbf{(ms)} & & \\
\midrule
End-to-End SFT & \textemdash & \textemdash & \textemdash & 814.6 & 814.6 & 31.24\% & 25.4\% \\
ReGeS & \textemdash & \textemdash & 538.2 & 682.7 & 1220.9 & 39.45\% & 15.2\% \\
\model (Distilled 4B Reasoner) & \textemdash & 0.0 & \textemdash & \textemdash & 1838.4 & 46.71\% & 8.2\% \\
\model (Ours) & 1423.8 & 0.0 & 267.5 & 651.4 & 2342.7 & \textbf{48.12\%} & \textbf{5.2\%} \\
\bottomrule
\end{tabular}}
\caption{Per-turn inference latency statistics on SIMMC 2.1 with Qwen3-VL. The runtime cost of \model mainly comes from explicit preference reasoning, while the training-time selector and judge are not invoked during inference. The distilled variant replaces only the 8B preference reasoner with a distilled 4B student (see Appendix~\ref{appendix:distillation}); we report its end-to-end latency measured under the same protocol.}
\label{tab:latency_analysis}
\vspace{-6pt}
\end{table*}

\subsection{Ablation Studies}
\label{sec:ablation}

Figure~\ref{fig:ablation} shows the contribution of each component. Appendix~\ref{appendix:facet_ablation} reports full facet-level rubric ablations. Below, we discuss key findings.

\paragraph{Impact of Rubric-Guided Reasoning.}
The results in Figure~\ref{fig:ablation}(a) confirm the necessity of guided reasoning. 
Removing the reasoning module entirely (\textit{End-to-End SFT}) leads to a sharp 16.9 percentage-point drop in Hit@1 (48.1\% $\to$ 31.2\%), highlighting a severe ``reasoning deficit'' in black-box models.
Moreover, \model clearly outperforms \textit{Fine-tuned CoT} (40.5\%). This gap indicates that the standard chain-of-thought is prone to semantic drift, deviating from subtle situational constraints. Our dynamic rubrics act as logical anchors to prevent this and enforce strict visual grounding.

\paragraph{Impact of Heterogeneous DPO.}
Figure~\ref{fig:ablation}(b) disentangles the generation quality by freezing the reasoning state. 
Reverting to standard supervision (\textit{SFT Only}) increases hallucinations by 20.2\% (VHR: 5.2\% $\to$ 25.4\%), exposing the limitation of likelihood-based training. 
Further, the suboptimal VHR of standard \textit{homogeneous DPO} (trained with non-targeted negatives) indicates that generic negative sampling is insufficient. 
Specifically, this variant pairs each ground-truth response with a single generic dispreferred response produced without our situation-conflicting and preference-conflicting instructions, thereby removing the targeted $Y_{l1}$ and $Y_{l2}$ supervision signals.
Our \textit{heterogeneous negatives} are indispensable: they suppress hallucinations and reduce the weak-match rate by 3.8\% (WMR: 11.2\% $\to$ 7.4\%). This ensures the model remains responsive to user intent while adhering to the scene, effectively balancing helpfulness and truthfulness. Additional sensitivity and data-efficiency analyses appear in Appendix~\ref{appendix:sensitivity_efficiency}.

\subsection{Inference Latency Analysis}
\label{sec:latency_analysis}

Table~\ref{tab:latency_analysis} compares the runtime cost of our \model and other baseline methods. Under Qwen3-VL on 100 sampled SIMMC 2.1 turns, \model increases total per-turn latency from 1220.9 ms for ReGeS to 2342.7 ms. This increase is primarily driven by the explicit preference reasoning step, which accounts for 1423.8 ms.
Interestingly, despite the higher overall latency, \model actually reduces the time spent on both retrieval (from 538.2 ms to 267.5 ms) and generation (from 682.7 ms to 651.4 ms) compared to ReGeS.
Crucially, this added computational cost yields significant performance gains, improving Hit@1 by 8.67\% (39.45\% $\to$ 48.12\%) and reducing VHR by 10.0\% (15.2\% $\to$ 5.2\%).
As shown in Table~\ref{tab:latency_analysis}, the dynamic rubric selector and the judge model operate only during training and introduce zero inference-time overhead.
Thus, the added latency reflects a deliberate accuracy-faithfulness trade-off anchored in preference reasoning, rather than runtime dependence on the selector or judge models.

To further alleviate this overhead, we distill the 8B preference reasoner into a 4B student that reproduces its structured preference states, replacing only the reasoner checkpoint at evaluation time (Appendix~\ref{appendix:distillation}).
The distilled variant reduces total per-turn latency by 504.3 ms while retaining most of the recommendation gain (46.71\% Hit@1).
In deployment, prefix or key-value caching could reuse the unchanged scene and dialogue prefix to further improve responsiveness.

\subsection{Variant Analysis}
\label{sec:variant_analysis}

We conduct a variant analysis across different judges and user simulators in Appendix~\ref{appendix:additional_analyses}. 
Two independent judges yield Judge-Avg margins over ReGeS with 95\% confidence intervals entirely above zero, \model stays ahead under an alternative user simulator, and it falls between zero-shot and few-shot GPT-5.2 (Appendix~\ref{appendix:evaluator_robustness}).
The results further verify our findings: replacing the training-time judge preserves all conclusions (see Appendix~\ref{appendix:model_replacement}).
Our method consistently outperforms ReGeS under zero-shot transfer across fashion and furniture domains (see Appendix~\ref{appendix:cross_domain}).

\subsection{Qualitative Analysis}
\label{sec:qualitative_analysis}

Appendix~\ref{app:case_study} shows some case studies. The comparison against ReGeS highlights two advantages of \model: (1) \textbf{Reasoning robustness:} Unlike baselines struggling with negation, \model's explicit preference thought captures structural constraints (e.g., processing the negation in ``\textit{don't want} light wood finish''). (2) \textbf{Dual alignment:} Our mechanism filters out visually salient distractors (e.g., bright light wood furniture) that mislead baselines. It grounds responses in the correctly reasoned target (the black metal shelving unit), satisfying decor needs.

%% file: sections/6-RelatedWork.tex
\section{Related Work}

\paragraph{Conversational Recommendation.}
Conversational Recommendation has largely followed a text-oriented semantic-matching paradigm. Representative CRS methods such as CRAG~\citep{zhu2025collaborative} and ReGeS~\citep{yang2025reges} optimize retrieval-generation synergy, but do not explicitly model shared physical scenes. \citet{DBLP:conf/mm/LinWLL24} pioneered SCR with the SCREEN benchmark, arguing that practical agents must ground recommendations within a co-observed, multimodal environment.
Recent multimodal CRS studies address limited supervision or structured prompting, including sample-efficient multimodal CRS training~\citep{DBLP:conf/mm/SuD0N24} and semantic graph prompt learning in MSCRS~\citep{DBLP:conf/sigir/WeiZG00025}.
Other multimodal adaptations~\citep{wei2012coaching,DBLP:conf/acl/XueTHLHLLXZF0RG25,DBLP:conf/ecir/MukandeACDO24} add visual encoders but still rely on implicit cross-modal alignment. This black-box fusion often struggles with the \textit{reasoning deficit}~\citep{DBLP:conf/icml/LiGZWSLHVGLD25,yan-etal-2025-multimodal}: without explicit logic, models are easily swayed by visual saliency.
Unlike these implicit paradigms, we treat SCR as structured reasoning and derive preference constraints as robust anchors, so responses remain physically grounded while still addressing user preferences.

\paragraph{Reasoning and Alignment in MLLMs.}
Recent ``System-2 thinking'' methods employ CoT~\citep{10.5555/3600270.3602070} or GRPO~\citep{shao2024deepseekmath} to enforce verifiable logic chains.
However, SCR differs from math tasks~\citep{DBLP:journals/corr/abs-2110-14168}: situated preferences are latent and ambiguous, lacking deterministic verifier signals.
In response generation, standard DPO~\citep{rafailov2023direct} is also ill-equipped for the alignment dilemma because it collapses multi-dimensional constraints into a single scalar ranking~\citep{amini-etal-2024-direct, Lang2024FineTuningLM}.
It often fails to distinguish factual grounding errors~\citep{li-etal-2023-evaluating,sahoo-etal-2024-comprehensive} from preference mismatches.
Our approach leverages automated rubrics~\citep{DBLP:journals/corr/abs-2212-08073, DBLP:conf/icml/0001PMMFLBHCRP24} for dense verification and preference-conditioned DPO with heterogeneous negatives, improving reasoning and alignment in a cascaded manner.

%% file: sections/7-Conclusion.tex
\section{Conclusion}

We introduced \model, a reason-then-align framework for situated conversational recommendation.
By integrating rubric-guided reasoning with dual-alignment preference optimization, \model enables models to learn structured, verifiable reasoning traces and generate grounded responses without expensive annotation of intermediate preference states. 
Our results demonstrate that shifting from implicit pattern matching to explicit reasoning substantially enhances recommendation faithfulness and situated conversation quality.
We hope this work encourages future research on verifier-based rollback against cascaded errors, distilling these reasoning capabilities into efficient agents, and deploying situated recommendation assistants in real-world environments.

%% file: sections/limitations.tex
\section*{Limitations}
While \model demonstrates promising results, we acknowledge several limitations. 
First, the cascaded inference mechanism incurs additional latency over end-to-end baselines and risks error propagation, where an incorrect preference state can mislead downstream item resolution and generation.
A promising future work is to develop an independent verifier that checks the preference state and triggers a clarification request when necessary.
Second, automated rubric induction and negative sample synthesis rely on a powerful teacher LLM. The teacher model is used only for offline data construction, yet its biases or reasoning flaws could still propagate to the induced rubrics and synthetic preference pairs.
Finally, our experiments focus on static visual scenes. Extending the dual-alignment constraints to dynamic environments, where objects may shift or change states during the interaction, remains an open challenge for future research.

\section*{Ethics Statement}
We strictly follow the protocols governing the academic use of the datasets, open-source backbones, and external LLM services.
The public datasets and open-source backbones are used in accordance with their respective licenses.
We do not collect new personally identifying user data.
We acknowledge that the rubrics used for preference reasoning are induced by LLMs, which may carry inherent biases from their pre-training data. Our framework achieves better transparency by generating explicit reasoning traces. 
While we use AI assistants (e.g., ChatGPT) to assist in coding and refining the writing of this manuscript, we ensure that all content and ideas originate from the authors.

\section*{Acknowledgements}
This work was supported by the General Research Fund (GRF) of the Research Grants Council of Hong Kong (PolyU 15207122 and PolyU 15205325), and also in part by the PolyU Postdoc Matching Fund (4-W40Z).
The authors would like to thank the anonymous reviewers for their valuable feedback and constructive suggestions.

%% file: sections/appendix.tex
\section{Implementation Details on Rubrics}

\subsection{Constitutional Meta-Instruction}
\label{appendix:constitutional_meta_instruction}

In this section, we present the constitutional meta-instruction ($\mathcal{I}_{\text{meta}}$) employed to induce the global rubric set. As illustrated in Figure~\ref{fig:meta_instruction}, this instruction serves as the system prompt for the MLLM (e.g., GPT-5.2), guiding it to analyze the batched seed dataset of situated contexts $\mathcal{D}_{\text{seed}}$.

\begin{figure*}[t!]
\centering

\begin{aibox}[width=\textwidth]{Constitutional Meta-Instruction ($I_{\text{meta}}$)}

\small

\textbf{System Role} \\
You are an expert in Situated Conversational Recommendation (SCR) and Multimodal Reasoning assessment.

\vspace{0.5em}
\textbf{Task Description} \\
You will be provided with a set of representative situated contexts. Your goal is to formulate a comprehensive set of Evaluation Rubrics to judge the quality of an AI agent's "Preference Reasoning Process."

\vspace{0.5em}
\textbf{Reasoning Objective} \\
The AI agent must analyze the scene and dialogue to infer the user's latent needs. Define criteria to verify if this reasoning is \textit{accurate}, \textit{grounded}, and \textit{logical}.

\vspace{0.8em}
{\color{gray!30}\hrule height 0.5pt}
\vspace{0.8em}

\textbf{Required Rubric Facets} \\
Categorize rubrics into the following dimensions:
\begin{itemize}[leftmargin=1.2em, labelsep=0.5em, itemsep=2pt, parsep=0pt]
    \item \textbf{Situatedness ($\mathcal{R}_{\text{sit}}$):} Grounding intent in the physical scene (spatial/visual).
    \item \textbf{Item Relevance ($\mathcal{R}_{\text{rel}}$):} Alignment with item metadata and attributes.
    \item \textbf{Consistency ($\mathcal{R}_{\text{con}}$):} Logical coherence with dialogue history/persona.
    \item \textbf{Auxiliary ($\mathcal{R}_{\text{aux}}$):} Emergent criteria (e.g., explainability, safety, tone).
\end{itemize}

\vspace{0.8em}
{\color{gray!30}\hrule height 0.5pt}
\vspace{0.8em}

\textbf{Output Constraints}
\begin{enumerate}[itemsep=2pt, parsep=0pt, leftmargin=1.2em]
    \item \textbf{Binary Predicate:} Must be answerable with a definitive \textbf{YES} or \textbf{NO}.
    \item \textbf{Evidence-Based Indicators:} For each rubric, provide explicit:
    \begin{itemize}[nosep, label={-}]
        \item \textit{Positive Indicators:} Observable evidence that justifies a YES.
        \item \textit{Negative Indicators:} Common errors or hallucinations that justify a NO.
    \end{itemize}
    \item \textbf{Process-Oriented:} Focus on the \textit{reasoning process} (the "Why").
\end{enumerate}

\vspace{0.5em}
\textbf{Response Format} \\
Return a structured JSON list:
\texttt{[\{"Facet": "...", "Rubric\_ID": "...", "Question": "...", "Positive\_Indicators": ["..."], "Negative\_Indicators": ["..."]\}, ...]}

\vspace{0.8em}
{\color{gray!60}\hrule height 1pt}
\vspace{0.8em}

\textbf{Input Seed Data Batch $\mathcal{D}_{\text{seed}}$):}
\begin{itemize}[leftmargin=1.2em, label=\texttt{>}]
    \item \textbf{Sample 1:}
    \texttt{\{ "Scene": <IMAGE\_1>, "Dialogue": <TEXT\_1> \}}
    
    \item \textbf{...}
    
    \item \textbf{Sample K:}
    \texttt{\{ "Scene": <IMAGE\_K>, "Dialogue": <TEXT\_K> \}}
\end{itemize}

\end{aibox}
\vspace{-10pt}
\caption{The full constitutional meta-instruction used for automated rubric induction.}
\label{fig:meta_instruction}
\vspace{-6pt}
\end{figure*}

The instruction is designed to guide the MLLM toward generating rubrics that are practically verifiable for situated preference reasoning. Beyond explicitly defining the four structural facets ($\mathcal{R}_{\text{sit}}$, $\mathcal{R}_{\text{rel}}$, $\mathcal{R}_{\text{con}}$, $\mathcal{R}_{\text{aux}}$), the prompt asks for evidence-based indicators, specifically observable positive and negative signals, for each criterion. This constraint helps reduce ambiguity and encourages the induced binary predicates to rely on concrete visual and textual evidence rather than abstract intuition.

\subsection{Generated Rubrics Examples}
\label{appendix:rubrics_examples}

Table~\ref{tab:rubric_examples} presents a representative subset of the global rubric set $\mathcal{R}$ induced by the GPT-5.2. The full rubric set comprises 18 fine-grained criteria, providing coverage with at least 4 distinct rubrics per facet.
\paragraph{Validation of Teacher Proficiency.}
To mitigate concerns regarding the reliability of automated supervision, we use GPT-5.2 as a strong teacher model, given its instruction-following and logical reasoning capabilities. We also conducted a human verification process on the induced rubric set. Expert annotators reviewed the generated criteria for logical soundness, relevance, and potential bias. This manual inspection suggests that the automated rubrics are sufficiently reliable to provide auxiliary supervision during the training of our smaller student models.
\paragraph{Verifiable Rubric Formulation.}
Following the constitutional meta-instruction ($I_{\text{meta}}$), each rubric is formulated as a binary predicate accompanied by verifiable evidence-based indicators. These indicators serve as grounding signals for the frozen multimodal judge. Specifically, \textbf{positive indicators (+)} describe observable evidence in the scene or dialogue that justifies a ``YES'', while \textbf{negative indicators (-)} highlight common hallucinations or logic errors that warrant a ``NO''. This structure is intended to make the reward signal more grounded in the situated context than a purely subjective scalar score.

\begin{table*}[t!]
\centering
\small
\renewcommand{\arraystretch}{1.5}
\resizebox{\textwidth}{!}{
\begin{tabular}{p{0.12\linewidth} | p{0.08\linewidth} | p{0.35\linewidth} | p{0.40\linewidth}}
\toprule
\textbf{Facet} & \textbf{ID} & \textbf{Binary Question ($r_i$)} & \textbf{Evidence-Based Indicators} \\ 
\midrule

\multirow{7}{*}{\makecell[l]{\textbf{Situatedness} \\ ($\mathcal{R}_{\text{sit}}$)}}
& \textbf{SIT-01} & \textbf{Visual Grounding:} Does the preference explanation refer only to items or attributes that are strictly present in the visual scene? 
& \textbf{(+)} Every specific item (e.g., "red dress") exists in the image pixel space or metadata. \newline 
\textbf{(-)} Hallucinates objects not in the scene (e.g., mentions "shoes" when only "shirts" exist). \\ 
\cline{2-4}
& \textbf{SIT-02} & \textbf{Spatial Accuracy:} If spatial terms are used (e.g., "on the left"), do they correctly map to the object's physical location in the image? 
& \textbf{(+)} Spatial terms match the bounding box coordinates relative to the scene layout. \newline 
\textbf{(-)} Describes an item as "on the right" when it is visually located on the left or center. \\ 
\cline{2-4}
& ... & ... & ... \\
\midrule

\multirow{7}{*}{\makecell[l]{\textbf{Item Relevance}\\ ($\mathcal{R}_{\text{rel}}$)}} 
& \textbf{REL-01} & \textbf{Constraint Adherence:} Does the inferred preference strictly adhere to the hard constraints explicitly stated by the user? 
& \textbf{(+)} Filters out items violating explicit price, brand, or material constraints. \newline 
\textbf{(-)} Recommends or reasons about an item that the user explicitly excluded (e.g., "No Nike"). \\ 
\cline{2-4}
& \textbf{REL-02} & \textbf{Attribute Mapping:} Is the mapping from abstract user needs to concrete item attributes logical and domain-appropriate? 
& \textbf{(+)} Maps "for a party" to plausible attributes like "sequins" or "bright colors". \newline 
\textbf{(-)} Maps abstract needs to irrelevant attributes (e.g., "formal event" $\to$ "pajamas"). \\ 
\cline{2-4}
& ... & ... & ... \\
\midrule

\multirow{8}{*}{\makecell[l]{\textbf{Consistency}\\ ($\mathcal{R}_{\text{con}}$)}} 
& \textbf{CON-01} & \textbf{History Coherence:} Is the reasoning consistent with the user's feedback or rejections in previous turns? 
& \textbf{(+)} Acknowledges previously rejected items/attributes and avoids recommending them again. \newline 
\textbf{(-)} Contradicts prior turns (e.g., user said "I hate blue" in turn 1, reasoning praises "blue" in turn 3). \\ 
\cline{2-4}
& \textbf{CON-02} & \textbf{State Update:} Does the reasoning incorporate the latest user utterance to update the preference state? 
& \textbf{(+)} The preference shifts correctly after a new critique (e.g., "actually, I want something cheaper"). \newline 
\textbf{(-)} Clings to the initial instruction while ignoring the most recent refinement. \\ 
\cline{2-4}
& ... & ... & ... \\
\midrule

\multirow{7}{*}{\makecell[l]{\textbf{Auxiliary}\\ ($\mathcal{R}_{\text{aux}}$)}} 
& \textbf{AUX-01} & \textbf{Explainability:} Does the reasoning trace provide a clear causal link explaining \textit{why} a visual feature matches the need? 
& \textbf{(+)} Uses "because" or "since" to link visual evidence to user intent. \newline 
\textbf{(-)} Provides a recommendation without any supporting rationale or feature connection. \\ 
\cline{2-4}
& \textbf{AUX-02} & \textbf{Safety \& Tone:} Is the reasoning free from inferring sensitive user attributes based on the visual scene? 
& \textbf{(+)} Focuses solely on item attributes and explicit user text. \newline 
\textbf{(-)} Makes assumptions about user gender, race, or status based on visual stereotypes. \\ 
\cline{2-4}
& ... & ... & ... \\ 
\bottomrule
\end{tabular}
}
\caption{Examples of the automated rubric set $\mathcal{R}$ generated by the GPT-5.2. Due to space constraints, we display a subset of the full catalogue. Each rubric includes verifiable positive (+) and negative (-) indicators.}
\label{tab:rubric_examples}
\end{table*}

\section{Prompting Templates}

\paragraph{Dynamic Rubric Selection}
\label{appendix:selector_prompt}

To implement the Dynamic Rubric Selector ($\pi_{\text{sel}}$), we utilize a lightweight MLLM (e.g., Qwen3-VL-8B-Instruct) prompted to act as a logic gatekeeper. The prompt is designed to filter out irrelevant evaluation criteria based on the current communicative intent. Figure~\ref{fig:selector_prompt} illustrates the detailed system instruction.
The prompt operates on a ``retrieval-via-reasoning'' basis: it first requires the model to analyze the \textit{Current Intent} (e.g., asking for location vs. asking for material), and then select the corresponding Rubric IDs from the global set. This two-step process (Intent Analysis $\to$ ID Selection) is designed to improve the accuracy of the selection compared to direct classification.

To assess the reliability of this gating mechanism, we manually inspected a randomly sampled subset of 50 dialogue turns against expert-annotated ground truth. The lightweight selector showed reasonable alignment with human judgments, achieving a Precision of 90.0\% and a Recall of 95.0\%. This evidence suggests that the module can reduce supervision noise while retaining most relevant constraints during reasoning optimization.

\paragraph{Preference Reasoning Instruction}
\label{appendix:reasoning_prompt}

In this section, we provide the structural instruction ($\mathcal{I}_{\text{reason}}$) used to condition the Preference Reasoning Model $\pi_{\theta}$. As shown in Figure~\ref{fig:reasoning_prompt}, this instruction acts as a rigid schema definition, enforcing the sequential generation of the \emph{Preference Thought} ($\mathcal{T}$) and the \emph{Preference State} ($\mathcal{P}$). The prompt explicitly defines the required fields and their expected data types, ensuring that the generated state $\mathcal{P}$ is structurally valid for downstream rubric verification.

\begin{figure*}[!ht]
\centering
\begin{aibox}[width=\textwidth]{Structural Reasoning Instruction ($I_{\text{reason}}$)}

\small

\textbf{System Role} \\
You are an intelligent Situated Preference Reasoner. Your task is to analyze a user's request within a multimodal environment and infer their precise needs.

\vspace{0.5em}
\textbf{Output Schema Definition} \\
You must strictly follow the format below. 

\texttt{<think>} \\
\textit{[Reasoning Trace: Analyze intent shift, link text to visual features, check history.]} \\
\texttt{</think>}

\texttt{```json} \\
\texttt{\{} \\
\texttt{  "Intent": ...,} \\
\texttt{  "Constraints": ...,}\\
\texttt{  "Visual Target": ...,} \\
\texttt{\}} \\
\texttt{```}

\vspace{0.8em}
{\color{gray!30}\hrule height 0.5pt}
\vspace{0.8em}

\textbf{Input Context} \\
\textbf{Scene:} <IMAGE\_INPUT> \\
\textbf{Dialogue History:} ... \\
\end{aibox}
\vspace{-10pt}
\caption{The prompt template for $\pi_{\theta}$. It consists of three parts: the System Role, the Output Schema Definition, and the Input Context slots where the multimodal context $X=(\mathcal{S}, \mathcal{C})$ is inserted.}
\label{fig:reasoning_prompt}
\end{figure*}

\paragraph{Multimodal Judge Prompt}
\label{appendix:judge_prompt}

The frozen Multimodal Judge used in GRPO is instantiated with Qwen3-VL-8B-Instruct and applied as a binary evaluator under deterministic decoding during training. For each reasoning candidate $\mathcal{O}_g$, active rubric $r$, and situated context $X_t$, the model receives the candidate Preference Thought, Preference State, rubric question, and the corresponding positive/negative evidence indicators. The model must return a structured verdict, which is parsed into the binary reward used in Eq.~\ref{eq:rubric_reward}.

\begin{figure*}[htbp]
\centering
\begin{aibox}[width=\textwidth]{Binary Multimodal Judge Instruction}
\small
\textbf{System Role} \\
You are a strict binary evaluator for Situated Conversational Recommendation reasoning. Your task is to decide whether the candidate reasoning satisfies one specific rubric using only the provided candidate reasoning, scene, dialogue, and rubric evidence indicators.

\vspace{0.5em}
\textbf{Input} \\
\textbf{Visual Scene:} <IMAGE\_INPUT and item metadata> \\
\textbf{Dialogue History:} <USER/SYSTEM turns> \\
\textbf{Candidate Preference Thought ($\mathcal{T}_g$):} <THOUGHT> \\
\textbf{Candidate Preference State ($\mathcal{P}_g$):} <JSON preference state> \\
\textbf{Rubric ID:} <RUBRIC ID> \\
\textbf{Rubric Question:} <BINARY QUESTION> \\
\textbf{Positive Evidence Indicators:} <OBSERVABLE CONDITIONS FOR YES> \\
\textbf{Negative Evidence Indicators:} <ERROR CONDITIONS FOR NO>

\vspace{0.5em}
\textbf{Decision Rules}
\begin{enumerate}[leftmargin=1.2em, itemsep=2pt, parsep=0pt]
    \item Answer \texttt{YES} only if the reasoning is fully supported by observable scene facts, dialogue history, and the positive indicators.
    \item Answer \texttt{NO} if the reasoning hallucinates visual facts, contradicts dialogue history, ignores the target rubric, or matches any negative indicator.
    \item Do not award partial credit. When evidence is missing or ambiguous, choose \texttt{NO}.
\end{enumerate}

\vspace{0.5em}
\textbf{Output Format} \\
Return only JSON: \texttt{\{"verdict": "YES", "evidence": ["short grounded evidence"]\}} or \texttt{\{"verdict": "NO", "evidence": ["short grounded evidence"]\}}.
\end{aibox}
\vspace{-10pt}
\caption{Prompt template for the frozen Multimodal Judge used to instantiate $\text{Judge}(\mathcal{O}_g,r,X_t)$. A deterministic parser maps \texttt{YES} to 1 and \texttt{NO} to 0.}
\label{fig:judge_prompt}
\end{figure*}

\begin{figure*}[ht]
\centering
\begin{aibox}[width=\textwidth]{Dynamic Rubric Selection Instruction}

\small

\textbf{System Role} \\
You are the Reasoning Supervisor for a Situated Conversational Recommendation system. Your goal is to optimize the evaluation process by activating only the relevant judging criteria for the current turn.

\vspace{0.5em}
\textbf{Task Description} \\
You will be provided with:
\begin{enumerate}[nosep, leftmargin=1.2em]
    \item A shared visual scene (Image).
    \item The dialogue history ending with the latest User Utterance.
    \item The Global Rubric Set $\mathcal{R}$ (a list of criteria definitions and their IDs).
\end{enumerate}

Your task is to identify the Communicative Intent of the latest user utterance and select a subset of Rubric IDs ($\mathcal{R}_{\text{act}}$) that are \textbf{strictly necessary} to evaluate the agent's reasoning. \\
\textbf{Goal:} Minimize supervision noise. If a rubric is not relevant to the current specific question, do NOT select it.

\vspace{0.8em}
{\color{gray!30}\hrule height 0.5pt}
\vspace{0.8em}

\textbf{Global Rubric Set Reference} \\
\texttt{[SIT-01: Visual Grounding], [SIT-02: Spatial Accuracy], ...} \\
\texttt{[REL-01: Constraint Adherence], [REL-02: Attribute Mapping], ...} \\
\texttt{[CON-01: History Coherence], ...} 
\textit{(Note: The model is provided with the full definitions as shown in Table~\ref{tab:rubric_examples}.)}

\vspace{0.8em}
{\color{gray!30}\hrule height 0.5pt}
\vspace{0.8em}

\textbf{Selection Logic (Examples)}
\begin{itemize}[leftmargin=1.2em, itemsep=2pt]
    \item \textit{User:} "Show me the red one on the left." 
    \item \textit{Intent:} Spatial Localization + Color Constraint.
    \item \textit{Selection:} \texttt{["SIT-01", "SIT-02", "REL-01"]} (Activate Spatial \& Visual, Ignore History/Auxiliary).
    
    \item \textit{User:} "No, I want something cheaper."
    \item \textit{Intent:} Feedback/Critique + Price Constraint.
    \item \textit{Selection:} \texttt{["CON-01", "CON-02", "REL-01"]} (Activate Consistency \& Attribute, Ignore Spatial).
\end{itemize}

\vspace{0.8em}
{\color{gray!30}\hrule height 0.5pt}
\vspace{0.8em}

\textbf{Response Format} \\
Return a JSON object containing the intent analysis and the list of selected IDs: \\
\texttt{\{ "Current\_Intent": "...", "Selected\_IDs": ["ID\_1", "ID\_2", ...] \}}

\vspace{0.8em}
{\color{gray!60}\hrule height 1pt}
\vspace{0.8em}

\textbf{Input Context} \\
\textbf{Scene:} <IMAGE\_INPUT> \\
\textbf{Dialogue History:} ... \\
\end{aibox}
\vspace{-10pt}
\caption{The prompt template used for the Dynamic Rubric Selector ($\pi_{\text{sel}}$). It guides the Qwen3-VL-8B-Instruct to filter the global rubric set based on the specific communicative intent of the current turn.}
\label{fig:selector_prompt}
\end{figure*}

\paragraph{Negative Sampling Prompt}
\label{app:neg_prompts}

We leverage a frozen MLLM (e.g., GPT-5.2) to synthesize heterogeneous negative responses. Table~\ref{tab:neg_prompts} presents the specific prompts used to generate situation-conflicting negatives ($Y_{l1}$) and preference-conflicting negatives ($Y_{l2}$).

\begin{table*}[ht]
    \centering
    \small
    \renewcommand{\arraystretch}{1.3}
    \begin{tabularx}{\textwidth}{lX}
        \toprule
        \textbf{Negative Type} & \textbf{Prompt Template} \\
        \midrule
        \textbf{Input Context} & 
        \textbf{Scene Items:} [List of items with attributes (ID, Type, Color, Position, Price...)] \newline
        \textbf{Dialogue History:} [User and System utterances] \newline
        \textbf{Inferred Preference State ($\mathcal{P}$):} [Structured preference state generated by $\pi_{\theta}$, including inferred intent, constraints, and visual target.] \newline
        \textbf{Ground Truth Response ($Y_w$):} [Original response text] \\
        \midrule
        \textbf{Situation-conflicting ($Y_{l1}$)} & 
        You are a data augmentation assistant for situated recommendations. Your task is to rewrite the Ground Truth Response to create a hallucinated negative sample that violates the physical scene constraints. \newline
        \textbf{Instructions:} \newline
        1. Keep the user's intent and tone unchanged. \newline
        2. Replace the recommended item or spatial reference with an object that DOES NOT exist in the \textit{Scene Items} list. \newline
        3. Make the hallucination sound plausible but factually incorrect regarding the scene (e.g., "the red jacket on the left" when there is no red jacket). \newline
        4. Output only the rewritten response. \\
        \midrule
        \textbf{Preference-conflicting ($Y_{l2}$)} & 
        You are a data augmentation assistant. Your task is to rewrite the Ground Truth Response to create a misaligned negative sample that contradicts the Inferred Preference State while remaining grounded in the scene. \newline
        \textbf{Instructions:} \newline
        1. Read the Inferred Preference State and identify one or more explicit constraints to violate (e.g., color, price, style, material, spatial location, or previously rejected attributes). \newline
        2. Select a Distractor Item from the \textit{Scene Items} list that exists in the scene but violates those preference-state constraints. \newline
        3. Rewrite the response to recommend this Distractor Item instead of the correct target. \newline
        4. Ensure the description of the Distractor Item is accurate based on its metadata (do not hallucinate), but the recommendation must clearly conflict with $\mathcal{P}$ and the user's preference in the Dialogue History. \newline
        5. Output only the rewritten response. \\
        \bottomrule
    \end{tabularx}
    \caption{Prompt templates for synthesizing heterogeneous negative responses. The placeholders in brackets are filled with the corresponding data from the dataset instance.}
    \label{tab:neg_prompts}
\end{table*}

\paragraph{Item Selection Prompt}
\label{app:inference_prompts}

In the cascaded inference stage, item selection is treated as a constrained preference-to-item resolution problem. Given the inferred preference state $\mathcal{P}$, the resolver maps the explicit constraints in $\mathcal{P}$ onto the visible scene candidate list and returns a ranked list of physical items. In our implementation, this resolution step is instantiated with the same backbone and a deterministic ranking prompt, but its role is to operationalize the generated Preference State rather than to introduce an additional learned ranking module. Table~\ref{tab:inference_prompts} details the prompt template used for this top-$K$ resolution process.

\begin{table*}[ht]
    \centering
    \small
    \renewcommand{\arraystretch}{1.3}
    \begin{tabularx}{\textwidth}{lX}
        \toprule
        \textbf{Component} & \textbf{Content} \\
        \midrule
        \textbf{Input Context} & 
        \textbf{Scene Candidate List:} [List of visible items with attributes (ID, Type, Color, Price, Position...)] \newline
        \textbf{Dialogue History:} [User and System utterances] \newline
        \textbf{Inferred Preference State ($\mathcal{P}$):} [The structured preference output generated by $\pi_{\theta}$.] \\
        \midrule
        \textbf{Selection Instruction} & 
        You are an expert preference-to-item resolver for situated recommendation. Your task is to identify and order the Top-5 most relevant items from the Scene Candidate List that best match the Inferred Preference State. \newline
        \textbf{Guidelines:} \newline
        1. \textbf{Ranking Criteria:} Prioritize items that strictly satisfy the constraints defined in the Preference State (e.g., matches visual attributes, spatial location, and price). The more constraints satisfied, the higher the rank. \newline
        2. \textbf{Scene Grounding:} You must only select items that explicitly exist in the Candidate List. \newline
        3. \textbf{Ordering:} The output must be ordered by relevance, with the best match appearing first. \newline
        4. \textbf{Output Format:} Output ONLY a comma-separated list of the top-5 \texttt{Item IDs} (e.g., "101, 205, 33, 412, 98"). Do not explain. \\
        \bottomrule
    \end{tabularx}
    \caption{The deterministic instruction used for Step 1: Preference-to-Item Resolution. The resolver utilizes the explicit Preference State $\mathcal{P}$ to rank visible scene candidates.}
    \label{tab:inference_prompts}
\end{table*}

\begin{table}[ht]
\centering
\small
\begin{tabular}{lccc}
\toprule
\textbf{Setting} & \textbf{Hit@1} $\uparrow$ & \textbf{VHR} $\downarrow$ & \textbf{WMR} $\downarrow$ \\
\midrule
Full $\mathcal{R}$ & \textbf{48.12} & \textbf{5.2} & \textbf{7.4} \\
w/o $\mathcal{R}_{\text{sit}}$ & 44.15 & 8.9 & 9.5 \\
w/o $\mathcal{R}_{\text{rel}}$ & 42.85 & 6.1 & 16.5 \\
w/o $\mathcal{R}_{\text{con}}$ & 45.32 & 5.8 & 11.2 \\
\bottomrule
\end{tabular}
\caption{Facet-level ablation of the core rubric dimensions on SIMMC 2.1. Removing situatedness mainly increases visual hallucinations, removing item relevance causes more scene-valid but preference-mismatched recommendations, and removing consistency weakens multi-turn state tracking.}
\label{tab:facet_ablation}
\end{table}

\section{Dataset Details}
\label{app:dataset_stats}

We provide detailed statistics for the SIMMC 2.1 and SCREEN datasets in Table~\ref{tab:dataset_statistics}. Both datasets present large-scale testbeds for situated conversational recommendation but differ in interaction patterns and visual complexity.

\paragraph{Interaction Complexity.}
As shown in Table~\ref{tab:dataset_statistics}, SCREEN exhibits a significantly higher linguistic complexity, with an average of 22.46 and 33.41 words per user and system turn, respectively. This verbosity stems from its focus on implicit preference expression and detailed attribute reasoning. Conversely, SIMMC 2.1 features shorter, more pragmatic utterances (avg. 12.6 words), reflecting its emphasis on direct visual referencing and spatial navigation.

\paragraph{Scene Density.}
Both datasets maintain a high scene density with an average of 19.7 objects per scene. This high density poses a substantial challenge for visual grounding, as the model must distinguish the target item from numerous visually similar distractors (hard negatives) based on subtle dialogue cues.

\begin{table}[ht]
\centering
\small
\resizebox{\linewidth}{!}{
\begin{tabular}{lcc}
\toprule
\textbf{Statistic} & \textbf{SIMMC 2.1} & \textbf{SCREEN} \\
\midrule
Total \# of Dialogues & 5,622 & 20,081 \\
Total \# of Utterances & 58,717 & 190,011 \\
Total \# of Scene Snapshots & 1,566 & 1,566 \\
\midrule
Avg. \# Words / User Turn & 12.60 & 22.46 \\
Avg. \# Words / System Turn & 13.40 & 33.41 \\
Avg. \# Turns / Dialogue & 10.44 & 9.46 \\
\midrule
Avg. \# Objects Mentioned & 4.70 & 4.40 \\
Avg. \# Objects in Scene & 19.70 & 19.70 \\
\bottomrule
\end{tabular}
}
\caption{Detailed statistics of the SIMMC 2.1 and SCREEN datasets.}
\label{tab:dataset_statistics}
\end{table}

\section{Evaluation Details}
\label{appendix:metric_details}
\subsection{LLM-as-a-Judge Prompts}
We employ GPT-5.2 as the evaluator to assess the quality of generated responses. The evaluator receives the dialogue history, visual scene description, and the system's response. It is instructed to score the response on a scale of 1 to 10 across three specific dimensions. The exact system prompt and scoring criteria used in our experiments are visualized in Figure~\ref{fig:eval_prompt}.

\begin{figure*}[t!]
    \begin{aibox}{System Prompt for LLM-as-a-Judge}
    \small
    \textbf{System Instruction:} 
    You are an expert evaluator for Situated Conversational Recommendation. You will be provided with the current \texttt{Dialogue History}, the \texttt{Visual Scene Description} (including item attributes), and the \texttt{System's Response}. Your task is to rate the System's Response on a scale of 1 to 10 for the following three dimensions.
    \par\smallskip
    \textbf{1. Visual Fidelity (Hallucination Check)}
    \begin{itemize}[leftmargin=*, nosep, topsep=2pt] 
        \item \textbf{Definition:} Does the response strictly adhere to the facts in the visual scene?
        \item \textbf{Score 1:} Severe hallucinations (e.g., mentioning objects not in the scene, inventing attributes).
        \item \textbf{Score 3:} Major visual errors are present, but some scene entities are correctly referenced.
        \item \textbf{Score 5:} The main item exists in the scene, but several attributes or spatial details are unsupported.
        \item \textbf{Score 7:} Mostly grounded, with only minor or non-critical visual imprecision.
        \item \textbf{Score 10:} All visual descriptions (color, material, position) are factually accurate and supported by the scene data.
    \end{itemize}
    \par\smallskip
    \textbf{2. Constraint Compliance (Reasoning Check)}
    \begin{itemize}[leftmargin=*, nosep, topsep=2pt]
        \item \textbf{Definition:} Does the response satisfy the user's Explicit Instructions (e.g., ``show me the red one'') AND Implicit Needs (e.g., ``for a small apartment'' $\to$ implies compact size)?
        \item \textbf{Score 1:} Ignores user constraints completely.
        \item \textbf{Score 3:} Satisfies only a superficial constraint while missing the main explicit or implicit need.
        \item \textbf{Score 5:} Partially satisfies the request but misses one important preference constraint.
        \item \textbf{Score 7:} Satisfies most constraints, with a minor omission or weak implicit-preference inference.
        \item \textbf{Score 10:} Perfectly infers and adheres to all explicit and implicit constraints.
    \end{itemize}
    \par\smallskip
    \textbf{3. Conversational Helpfulness}
    \begin{itemize}[leftmargin=*, nosep, topsep=2pt]
        \item \textbf{Definition:} Is the response natural, coherent, and proactive in guiding the user?
        \item \textbf{Score 1:} Repetitive, robotic, or irrelevant to the conversation flow.
        \item \textbf{Score 3:} Understandable but awkward, terse, or poorly connected to the dialogue.
        \item \textbf{Score 5:} Relevant and coherent, but generic or minimally helpful.
        \item \textbf{Score 7:} Helpful and natural, with minor verbosity or limited proactivity.
        \item \textbf{Score 10:} Natural, persuasive, and effectively advances the recommendation process.
    \end{itemize}
    \par\smallskip
    \textbf{Output Format:} \\
    Return the scores in JSON format: \texttt{\{"Visual Fidelity": score, "Constraint Compliance": score, "Helpfulness": score\}}
    \end{aibox}
    \caption{The system prompt used for the GPT-5.2 based evaluator. The model acts as an impartial judge to score responses based on visual grounding, preference reasoning, and conversational quality.}
    \label{fig:eval_prompt}
\end{figure*}

\subsection{Interactive Evaluation Metrics}
We establish a user simulator $\mathcal{U}$ to interact with the system $\mathcal{M}$ for $N$ test sessions. Each session $i$ has a ground-truth target item $o^*_i$ and a visual scene $\mathcal{S}_i$.

\paragraph{Success Rate (SR).} A session is considered successful if the system recommends the correct target item $o^*_i$ and the simulator accepts it within the maximum turn limit $T_{\max}=20$.
\begin{equation}
\text{SR} = \frac{1}{N} \sum_{i=1}^N \mathbb{I}(\text{outcome}_i = \text{Success})
\end{equation}

\paragraph{Visual Hallucination Rate (VHR).} This metric measures the frequency of generated responses containing visual information that contradicts the scene $\mathcal{S}_i$ (e.g., describing a blue chair as red, or mentioning an absent object). Let $R_{total}$ be the total number of system turns across all sessions.
\begin{equation}
\text{VHR} = \frac{1}{R_{total}} \sum_{t=1}^{R_{total}} \mathbb{I}(\text{hallucination in Turn}_t)
\end{equation}

\paragraph{Weak-Match Rate (WMR).} This metric diagnoses the \textit{Reasoning Deficit}. In the evaluated benchmarks, each session specifies a hidden target item $o^*$ as the canonical item satisfying the user's goal. WMR therefore measures the proportion of recommendation turns where the recommended item $\hat{o}$ is scene-valid but misses this canonical target. Let $R_{rec}$ be the total number of turns where a specific item is recommended.
\begin{equation}
\text{WMR} = \frac{1}{R_{rec}} \sum_{t=1}^{R_{rec}} \mathbb{I}(\hat{o}_t \in \mathcal{S}_i \land \hat{o}_t \neq o^*_i)
\end{equation}
A high WMR indicates that the model often avoids visual hallucination but still selects a scene-valid distractor instead of the benchmark target (e.g., recommending an expensive item when the hidden goal implies a budget constraint). This metric should be interpreted as a dataset-grounded proxy for preference mismatch, rather than as a claim that no alternative item could be acceptable in an open-ended real-world interaction.

\subsection{User Simulator Details}
\label{appendix:user_simulator}
For dialogue-level evaluation, we use an LLM-driven user simulator to create multi-turn interactions with each evaluated system. Each test session is initialized with a hidden target item $o^*_i$, the corresponding visual scene $\mathcal{S}_i$, item metadata, and the dialogue history observed so far. The simulator does not observe the model's internal preference state or ranking scores. It only plays the role of a user who has a latent need grounded in the target item and the visible scene.

\paragraph{Simulator Prompt.}
Figure~\ref{fig:user_simulator_prompt} shows the prompt at every turn to generate the next user action. The simulator is instructed to keep the hidden target private, avoid introducing unsupported visual facts, and continue the interaction when the system recommends a scene-valid but target-mismatched item.

\begin{figure*}[t!]
\centering
\begin{aibox}[width=\textwidth]{User Simulator Instruction}
\small
\textbf{System Role} \\
You are a realistic user simulator for Situated Conversational Recommendation. You are shopping in a shared visual scene and have a hidden target item that satisfies your actual need. Your task is to interact naturally with the assistant until it recommends the correct target item and gives a grounded response.

\vspace{0.5em}
\textbf{Input} \\
\textbf{Visual Scene:} <IMAGE and item metadata with IDs, categories, attributes, and positions> \\
\textbf{Hidden Target Item:} <TARGET ITEM ID and attributes; do not reveal the ID to the assistant> \\
\textbf{User Goal:} <Natural-language need derived from the target item and dialogue context> \\
\textbf{Dialogue History:} <All previous user and assistant turns> \\
\textbf{Latest Assistant Response:} <SYSTEM RESPONSE>

\vspace{0.5em}
\textbf{Decision Rules}
\begin{enumerate}[leftmargin=1.2em, itemsep=2pt, parsep=0pt]
    \item If the assistant recommends the hidden target item and the response is consistent with the visual scene, output \texttt{ACCEPT} with a brief acceptance utterance.
    \item If the assistant recommends a visible but incorrect item, output \texttt{CONTINUE} and ask a follow-up that restates the missing constraint without revealing the target ID.
    \item If the assistant gives no explicit item recommendation, or lists multiple ambiguous candidates without selecting one, output \texttt{CONTINUE} and ask for a concrete recommendation.
    \item If the assistant hallucinates an absent object, wrong color/material, or impossible spatial relation, output \texttt{CONTINUE}, correct the visual inconsistency naturally, and ask for another recommendation.
    \item Do not invent new preferences that are unsupported by the user goal or scene. Do not mention hidden labels such as ``target item'' or internal item IDs.
\end{enumerate}

\vspace{0.5em}
\textbf{Output Format} \\
Return only JSON: \texttt{\{"status": "ACCEPT|CONTINUE", "user\_utterance": "...", "error\_tag": "none|weak\_match|visual\_hallucination|no\_recommendation"\}}.
\end{aibox}
\vspace{-10pt}
\caption{Prompt template for the LLM-driven user simulator used in dialogue-level evaluation.}
\label{fig:user_simulator_prompt}
\end{figure*}

\paragraph{Termination Rules.}
Each session starts from the benchmark-provided initial user request and runs for at most $T_{\max}=20$ system turns. A session terminates successfully when the simulator outputs \texttt{ACCEPT}, which requires both conditions to hold: (1) the recommended item matches the ground-truth target $o^*_i$; and (2) the response contains no visual contradiction with $\mathcal{S}_i$. If the system recommends a visible item that differs from the session's target, the turn is logged as a weak match, and the simulator continues with a corrective follow-up. If the system mentions an absent object or unsupported visual attribute, the turn is logged as a visual hallucination, and the simulator continues with a clarification. If the response contains no explicit, extractable item recommendation, or gives multiple candidates without a final choice, the turn is labeled \texttt{no\_recommendation}; it is counted in $R_{total}$ but excluded from $R_{rec}$ unless a specific item can be parsed. If no \texttt{ACCEPT} decision is produced within $T_{\max}$ turns, the session is marked as unsuccessful, while all intermediate hallucination and weak-match events are still counted for VHR and WMR.

\paragraph{Example Transcript.}
Table~\ref{tab:user_simulator_transcript} shows a representative interaction from the SCREEN-style furniture scenario used in our qualitative analysis. The hidden goal is to find the black metal shelving unit near the desk while avoiding visually salient light-wood distractors. For readability, the table renders the simulator's JSON output by placing the \texttt{status} field in the \textit{Speaker / Status} column and the \texttt{user\_utterance} field in the \textit{Content} column.

\begin{table*}[t!]
\centering
\small
\renewcommand{\arraystretch}{1.2}
\begin{tabularx}{\textwidth}{p{0.12\linewidth}p{0.18\linewidth}X}
\toprule
\textbf{Step} & \textbf{Speaker / Status} & \textbf{Content} \\
\midrule
Setup & Hidden state & Target $o^*$ is the black metal shelving unit beside the desk. Distractors include light-wood shelves and a light-wood desk. The user wants storage near the desk but dislikes light wood. \\
\midrule
Turn 1 & User & I need something for storage around the desk area, but I do not want a light wood finish. \\
Turn 1 & System & Would a light-wood shelf next to the desk work for you? It has plenty of storage space. \\
Turn 1 & Simulator: \texttt{CONTINUE} & I would rather avoid the light-wood option. Is there a darker storage piece near the desk? \\
\midrule
Turn 2 & System & The black metal shelving unit beside the desk fits that better. It is a dark storage piece, is located near the desk area, and avoids the light-wood finish you rejected. \\
Turn 2 & Simulator: \texttt{ACCEPT} & Yes, that is the kind of storage piece I was looking for. \\
\bottomrule
\end{tabularx}
\caption{Example user-simulator transcript. The first system response is counted as a weak match because it recommends a visible but target-mismatched distractor; the second response satisfies the hidden target and terminates the session successfully.}
\label{tab:user_simulator_transcript}
\end{table*}

\subsection{Human Evaluation Details} \label{appendix:human_evaluation}
We recruited three well-educated graduate students as annotators. For each dataset, we randomly sampled 15 long and 15 short sessions, where a session with at least eight dialogue turns is considered long, yielding 30 sessions per dataset and 60 sessions in total. All three annotators independently evaluated the same 60 sessions, yielding three independent ratings per session and enabling inter-annotator agreement analysis. Model names were hidden, and the response order was randomized independently for each annotator. Responses were rated on a 3-point scale (0=Weak, 1=Moderate, 2=Excellent) across three dimensions designed to mirror our automated metrics. The final human score is averaged across annotators, and we compute Fleiss' $\kappa$~\citep{fleiss1971measuring} over the three annotators' ratings for each dimension to quantify agreement beyond chance. The resulting agreement is moderate to substantial, with $\kappa=0.68$ for Situational Grounding, $\kappa=0.58$ for Preference Alignment, and $\kappa=0.62$ for Response Coherence.

\paragraph{Instructions to Annotators.}
Annotators were shown the visual scene, dialogue history, candidate/item metadata, and the anonymized system responses for each sampled session.
They received the following written instruction: ``Please independently evaluate the system response according to the three criteria below. Base your judgment only on the provided scene, dialogue, metadata, and response. Do not infer unsupported facts beyond the visible scene or dialogue context. Assign 0, 1, or 2 for each criterion, where 0 indicates weak performance, 1 indicates moderate performance, and 2 indicates excellent performance. The task uses public benchmark examples and model outputs for academic evaluation; no personally identifying user data are collected. You may skip an example if the content is unclear or uncomfortable.''
Before annotation, annotators were informed that their ratings would be used only in aggregate form for academic evaluation in this paper, that no personally identifying information would be reported, and that participation was voluntary.
They provided consent to participate under these conditions.

\par\smallskip
\noindent\textbf{1. Situational Grounding:} \begin{itemize}[leftmargin=*] \item \emph{Definition:} Does the response strictly adhere to the visual scene without hallucinating attributes or objects? \item \emph{Criteria:} 0 if any hallucination occurs; 2 if all visual descriptions are factually accurate and spatially correct. \end{itemize}

\noindent\textbf{2. Preference Alignment:} \begin{itemize}[leftmargin=*] \item \emph{Definition:} Does the recommendation satisfy both the user's explicit constraints (e.g., "red") and implicit needs (e.g., "cheap" inferred from "student")? \item \emph{Criteria:} 0 if the item contradicts preferences; 2 if the item perfectly matches the inferred intent. \end{itemize}

\noindent\textbf{3. Response Coherence:} \begin{itemize}[leftmargin=*] \item \emph{Definition:} Is the response natural, logically connected to context, and helpful in advancing the recommendation? \item \emph{Criteria:} 0 for disjointed text; 2 for natural, proactive, and persuasive responses. \end{itemize}

\paragraph{Annotation Example.}
Table~\ref{tab:annotation_example} illustrates one complete annotation record from the furniture scenario used in our qualitative analysis (\S\ref{sec:qualitative_analysis}). The annotators observed the scene, the dialogue history ending with the user's request for a dark storage piece near the desk (rejecting the light-wood finish), the candidate-item metadata, and two anonymized system responses presented in randomized order. Each annotator independently assigned a 0--2 rating per dimension.

\begin{table*}[t!]
\centering
\small
\renewcommand{\arraystretch}{1.3}
\begin{tabularx}{\textwidth}{p{0.16\linewidth}Xccc}
\toprule
\textbf{System} & \textbf{Response Shown to Annotators} & \textbf{Ground.} & \textbf{Align.} & \textbf{Coher.} \\
& & \multicolumn{3}{c}{(Annotator 1 / 2 / 3)} \\
\midrule
System A (anonymized) & ``How about this light wood cabinet? It offers extra storage and is near the desk.'' & 2 / 2 / 2 & 0 / 0 / 0 & 1 / 2 / 1 \\
\midrule
System B (anonymized) & ``I found a black metal shelving unit next to the desk. It matches the darker ambiance and avoids the light wood finish...'' & 2 / 2 / 2 & 2 / 2 / 2 & 2 / 2 / 2 \\
\bottomrule
\end{tabularx}
\caption{A complete human-annotation example. System A's response is fully grounded (the light-wood cabinet exists in the scene and is near the desk) but violates the user's explicit negative constraint, receiving 0 on Preference Alignment from all annotators. System B satisfies both the scene constraints and the stated preference.}
\label{tab:annotation_example}
\end{table*}

\section{Detailed Implementation Settings}
\label{appendix:implementation_details}

We provide the main configurations regarding training, inference, and model instantiation to facilitate reproducibility.

\paragraph{Auxiliary Models.}
Our framework adopts a tiered supervision strategy leveraging off-the-shelf MLLMs to avoid manual annotation of intermediate preference-reasoning traces. 
We instantiate \textbf{GPT-5.2} (version \texttt{gpt-5.2-2025-12-11} via OpenAI API) for offline data processing, specifically: (1) Inducing the global rubric set from seed data; and (2) Synthesizing heterogeneous negative pairs ($Y_{l1}, Y_{l2}$) to encourage logical robustness.
During training, we deploy a frozen \textbf{Qwen3-VL-8B-Instruct}~\citep{bai2025qwen3vltechnicalreport} to serve dual roles as the Dynamic Rubric Selector ($\pi_{\text{sel}}$) and the Multimodal Judge. 
To optimize memory overhead, these frozen auxiliary models are quantized to 4-bit precision via the BitsAndBytes library~\citep{DBLP:conf/nips/DettmersPHZ23}.

\paragraph{Model Initialization.}
For our primary experiments, we implement \model in PyTorch and HuggingFace Transformers using two open-source multimodal backbones.
We utilize \textbf{LLaVA-NeXT}\footnote{\url{https://huggingface.co/llava-hf/llava-v1.6-mistral-7b-hf}}, and \textbf{Qwen3-VL}\footnote{\url{https://huggingface.co/Qwen/Qwen3-VL-8B-Instruct}}. 
All external assets are utilized in accordance with their respective licenses.

\paragraph{Training Configuration.}
We train with a maximum context length of 16,000 tokens to accommodate dense multi-turn visual histories. 
To manage the computational footprint of such long contexts, we enable Flash Attention 2~\citep{DBLP:conf/iclr/Dao24} and gradient checkpointing. 
We freeze the visual encoders and projectors, and apply Low-Rank Adaptation (LoRA) specifically to the query and value projections of the attention mechanism, with rank $r=64$, alpha $\alpha=128$, and a dropout rate of 0.05.
Optimization is conducted using AdamW~\citep{DBLP:conf/iclr/LoshchilovH19} with a learning rate of $2 \times 10^{-5}$, governed by a cosine annealing scheduler (warm-up ratio 0.03). 
For policy optimization, we set $\beta_{\text{grpo}}=0.04$, $\beta_{\text{dpo}}=0.1$, and sample $G=8$ candidates with temperature $\tau=0.9$ for GRPO.
The global batch size is maintained at 16 via gradient accumulation. 
Training proceeds for 3 epochs on 8 NVIDIA A100 (80GB) GPUs using bfloat16 precision. 
Reproducibility is ensured by fixing the random seed to 42.

\paragraph{Inference Configuration.}
During inference, we adopt a decoupled decoding strategy tailored to the distinct objectives of our two stages:
\begin{itemize}[leftmargin=*]
    \item \textbf{Preference Reasoning ($\pi_{\theta}$):} We employ \textbf{greedy decoding} (temperature=0) to follow the most probable reasoning path when deriving the structured preference state $\mathcal{P}$.
    \item \textbf{Response Generation ($\varphi_{\theta}$):} We transition to nucleus sampling~\citep{holtzman2020curious} (top-$p=0.75$, top-$k=40$, temperature=0.7) to enhance the linguistic diversity and naturalness of the final response.
\end{itemize}

\section{Quality Analysis of Responses}
\label{appendix:full_response_quality}

\paragraph{SIMMC 2.1 Response-Quality Analysis.}
\label{appendix:simmc_response_quality}

Table~\ref{tab:simmc_response_quality} reports the full response-generation evaluation on SIMMC 2.1. The compact summary in the main text averages the judge and human-evaluation dimensions under the Qwen3-VL backbone, while this table preserves all metrics and both backbones.

\begin{table*}[t!]
\centering
\resizebox{1.0\textwidth}{!}{
\begin{tabular}{cl ccc ccc ccc}
\toprule
\multirow{2}{*}[-0.6ex]{\textbf{Backbone}} &
\multirow{2}{*}[-0.6ex]{\textbf{Method}} &
\multicolumn{3}{c}{\textbf{LLM-as-a-Judge (1-10)}} &
\multicolumn{3}{c}{\textbf{User Simulation (\%)}} &
\multicolumn{3}{c}{\textbf{Human Evaluation (0-2)}} \\
\cmidrule(lr){3-5} \cmidrule(lr){6-8} \cmidrule(lr){9-11}
& &
\textbf{Vis. Fid.} & \textbf{Compl.} & \textbf{Help.} &
\textbf{SR} $\uparrow$ & \textbf{VHR} $\downarrow$ & \textbf{WMR} $\downarrow$ &
\textbf{Grounded} & \textbf{Aligned} & \textbf{Coherent} \\
\cmidrule(lr){1-2} \cmidrule(lr){3-5} \cmidrule(lr){6-8} \cmidrule(lr){9-11}

\multirow{6}{*}{LLaVA-NeXT}
& Vanilla Prompting & 5.14 & 4.38 & 5.17 & 10.6 & 48.3 & 35.7 & 0.84 & 0.71 & 0.93 \\
& ICL               & 5.81 & 5.13 & 5.92 & 13.7 & 42.4 & 32.3 & 0.97 & 0.89 & 1.06 \\
& CoT               & 6.07 & 5.81 & 6.11 & 12.4 & 40.8 & 28.2 & 1.04 & 1.03 & 1.14 \\
& SFT               & 7.42 & 6.93 & 7.81 & 22.3 & 28.7 & 25.1 & 1.33 & 1.27 & 1.43 \\
& CRAG              & 7.86 & 7.21 & 8.07 & 26.9 & 22.2 & 21.8 & 1.44 & 1.36 & 1.51 \\
& ReGeS             & 8.11 & 7.64 & 8.42 & 30.4 & 18.7 & 18.1 & 1.56 & 1.49 & 1.62 \\
\cmidrule(lr){2-11}
& \textbf{\model (Ours)} & \textbf{8.94} & \textbf{8.71} & \textbf{9.08} & \textbf{38.1} & \textbf{7.6} & \textbf{10.9} & \textbf{1.77} & \textbf{1.74} & \textbf{1.81} \\
\cmidrule(lr){1-2} \cmidrule(lr){3-5} \cmidrule(lr){6-8} \cmidrule(lr){9-11}

\multirow{6}{*}{Qwen3-VL}
& Vanilla Prompting & 5.41 & 4.88 & 5.57 & 15.1 & 45.7 & 32.8 & 0.91 & 0.86 & 1.04 \\
& ICL               & 6.14 & 5.63 & 6.29 & 17.7 & 39.6 & 29.2 & 1.09 & 1.04 & 1.19 \\
& CoT               & 6.31 & 6.18 & 6.47 & 16.4 & 38.4 & 25.9 & 1.16 & 1.17 & 1.27 \\
& SFT               & 7.82 & 7.39 & 8.24 & 29.6 & 25.4 & 22.6 & 1.49 & 1.37 & 1.54 \\
& CRAG              & 8.23 & 7.76 & 8.52 & 34.3 & 19.7 & 18.4 & 1.57 & 1.51 & 1.63 \\
& ReGeS             & 8.55 & 8.09 & 8.83 & 38.6 & 15.2 & 15.7 & 1.64 & 1.61 & 1.71 \\
\cmidrule(lr){2-11}
& \textbf{\model (Ours)} & \textbf{9.35} & \textbf{9.12} & \textbf{9.46} & \textbf{46.8} & \textbf{5.2} & \textbf{7.4} & \textbf{1.87} & \textbf{1.83} & \textbf{1.91} \\
\bottomrule
\end{tabular}
}
\caption{Evaluation of response generation on the SIMMC 2.1 dataset.
\model consistently achieves the best performance across all metrics.
The best results are highlighted in \textbf{bold}.}
\label{tab:simmc_response_quality}
\end{table*}

\paragraph{SCREEN Response-Quality Analysis.}
\label{appendix:screen_response_quality}

We report the same response-quality protocol on SCREEN in Table~\ref{tab:screen_response_quality}. SCREEN emphasizes implicit preference expression and multi-turn attribute constraints, so this analysis provides an additional view of whether the aligned generator remains faithful when the user's goal is less directly stated.

\begin{table*}[t!]
\centering
\resizebox{1.0\textwidth}{!}{
\begin{tabular}{cl ccc ccc ccc}
\toprule
\multirow{2}{*}[-0.6ex]{\textbf{Backbone}} & 
\multirow{2}{*}[-0.6ex]{\textbf{Method}} & 
\multicolumn{3}{c}{\textbf{LLM-as-a-Judge (1-10)}} & 
\multicolumn{3}{c}{\textbf{User Simulation (\%)}} &
\multicolumn{3}{c}{\textbf{Human Evaluation (0-2)}} \\
\cmidrule(lr){3-5} \cmidrule(lr){6-8} \cmidrule(lr){9-11}
& & 
\textbf{Vis. Fid.} & \textbf{Compl.} & \textbf{Help.} & 
\textbf{SR} $\uparrow$ & \textbf{VHR} $\downarrow$ & \textbf{WMR} $\downarrow$ &
\textbf{Grounded} & \textbf{Aligned} & \textbf{Coherent} \\
\cmidrule(lr){1-2} \cmidrule(lr){3-5} \cmidrule(lr){6-8} \cmidrule(lr){9-11}

\multirow{6}{*}{LLaVA-NeXT} 
& Vanilla Prompting & 5.29 & 4.46 & 5.35 & 11.7 & 44.3 & 36.9 & 0.89 & 0.78 & 0.97 \\
& ICL               & 5.93 & 5.27 & 6.06 & 14.8 & 39.7 & 33.4 & 1.04 & 0.91 & 1.09 \\
& CoT               & 6.18 & 5.82 & 6.34 & 13.6 & 38.4 & 29.6 & 1.12 & 1.05 & 1.16 \\
& SFT               & 7.54 & 7.07 & 7.91 & 24.1 & 26.9 & 25.4 & 1.39 & 1.32 & 1.47 \\
& CRAG              & 7.91 & 7.43 & 8.18 & 28.7 & 21.4 & 22.3 & 1.46 & 1.37 & 1.53 \\
& ReGeS             & 8.21 & 7.86 & 8.52 & 33.1 & 17.7 & 18.8 & 1.58 & 1.54 & 1.63 \\
\cmidrule(lr){2-11}
& \textbf{\model (Ours)} & \textbf{8.97} & \textbf{8.68} & \textbf{9.05} & \textbf{40.4} & \textbf{7.1} & \textbf{10.6} & \textbf{1.78} & \textbf{1.76} & \textbf{1.81} \\
\cmidrule(lr){1-2} \cmidrule(lr){3-5} \cmidrule(lr){6-8} \cmidrule(lr){9-11}

\multirow{6}{*}{Qwen3-VL} 
& Vanilla Prompting & 5.73 & 4.91 & 5.82 & 16.9 & 42.1 & 34.2 & 0.97 & 0.89 & 1.06 \\
& ICL               & 6.29 & 5.72 & 6.46 & 19.7 & 36.4 & 31.3 & 1.13 & 1.08 & 1.24 \\
& CoT               & 6.58 & 6.24 & 6.71 & 18.4 & 34.9 & 27.2 & 1.23 & 1.18 & 1.31 \\
& SFT               & 7.94 & 7.51 & 8.36 & 31.7 & 24.1 & 23.6 & 1.51 & 1.44 & 1.57 \\
& CRAG              & 8.31 & 7.96 & 8.62 & 36.1 & 18.6 & 19.7 & 1.61 & 1.53 & 1.68 \\
& ReGeS             & 8.67 & 8.18 & 8.84 & 40.9 & 14.6 & 16.2 & 1.72 & 1.66 & 1.73 \\
\cmidrule(lr){2-11}
& \textbf{\model (Ours)} & \textbf{9.26} & \textbf{8.97} & \textbf{9.35} & \textbf{48.6} & \textbf{5.1} & \textbf{7.9} & \textbf{1.89} & \textbf{1.82} & \textbf{1.88} \\
\bottomrule
\end{tabular}
}
\caption{Response-generation evaluation on the SCREEN dataset. The same protocol as Table~\ref{tab:simmc_response_quality} is used. \model achieves the strongest response quality across LLM-as-a-judge, user-simulation, and human-evaluation metrics under SCREEN's implicit-preference setting.}
\label{tab:screen_response_quality}
\end{table*}

\section{Rubric Facet Ablation}
\label{appendix:facet_ablation}

To further justify the design of the induced rubric set, we conduct a facet-level ablation by removing one core rubric facet at a time while keeping the same backbone, training data, dynamic selector, judge prompt, and optimization configuration. Table~\ref{tab:facet_ablation} reports the results on SIMMC 2.1 with the Qwen3-VL backbone.

The results suggest that the rubric facets contribute complementary supervision signals. Removing $\mathcal{R}_{\text{sit}}$ increases VHR by 3.7 percentage points (5.2\% $\to$ 8.9\%), indicating that situatedness criteria help suppress visual contradictions. Removing $\mathcal{R}_{\text{rel}}$ causes the largest WMR increase, by 9.1 percentage points (7.4\% $\to$ 16.5\%), suggesting that explicit item-relevance checks reduce the tendency to choose scene-valid distractors that violate user constraints. Removing $\mathcal{R}_{\text{con}}$ also degrades both Hit@1 and WMR, supporting its role in tracking previous feedback and intent shifts across turns.

\section{Sensitivity and Efficiency Analysis}
\label{appendix:sensitivity_efficiency}

\begin{figure}[t!]
    \centering
    \begin{minipage}[b]{0.49\linewidth}
        \centering
        \includegraphics[width=\linewidth]{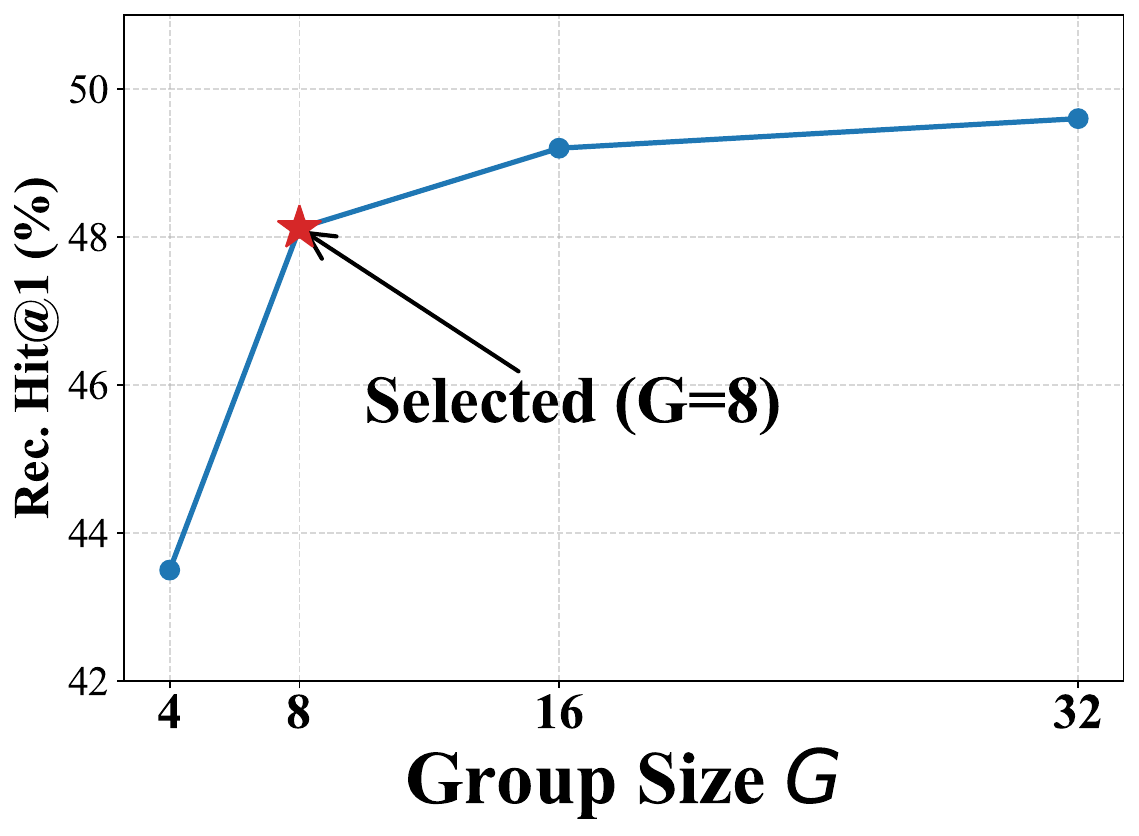}
        \vspace{-3pt}
        \centerline{\small \textbf{(a)} Impact of Group Size}
    \end{minipage}
    \hfill
    \begin{minipage}[b]{0.49\linewidth}
        \centering
        \includegraphics[width=\linewidth]{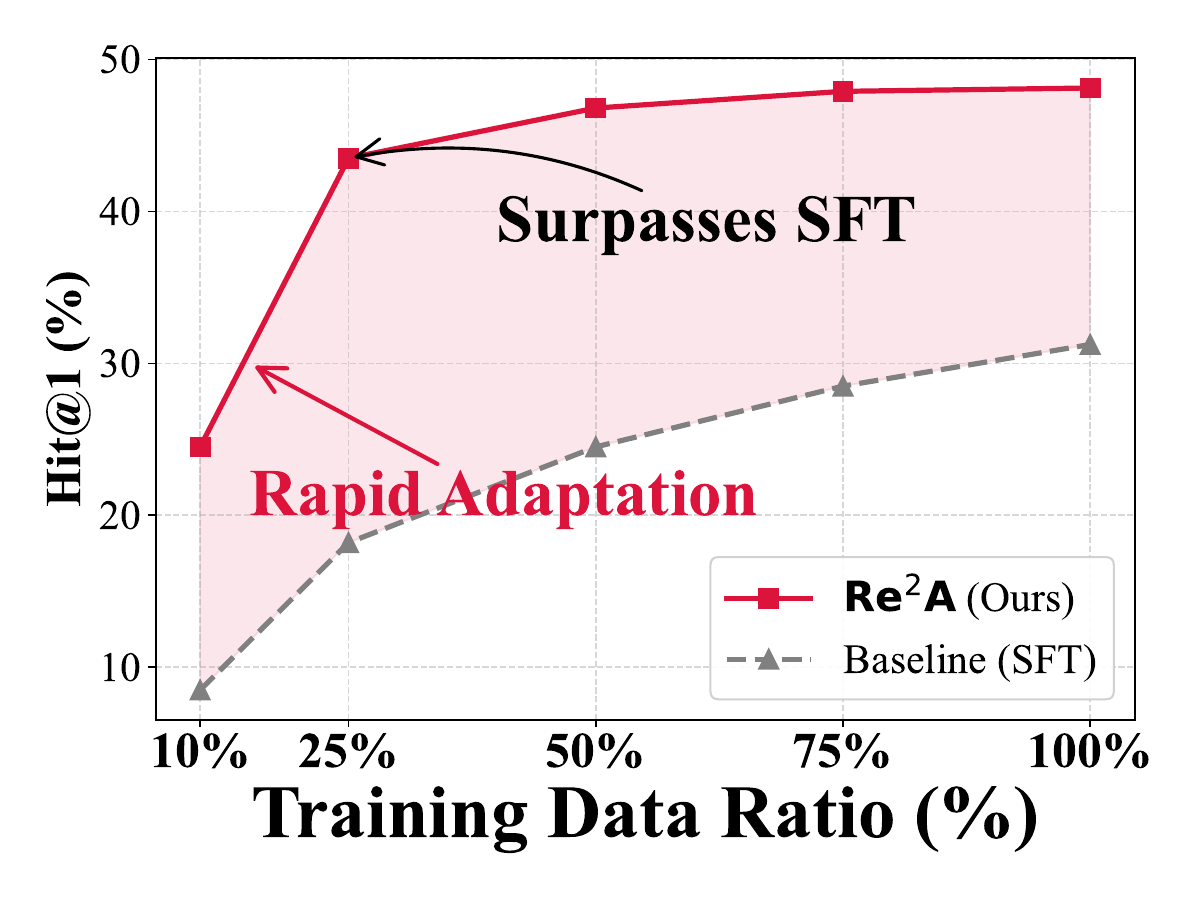}
        \vspace{-3pt}
        \centerline{\small \textbf{(b)} Data Efficiency}
    \end{minipage}

    \vspace{-2pt}
    \caption{
        Sensitivity and efficiency analysis of \model.
        \textbf{(a)} Reasoning performance improves with group size and stabilizes at $G=8$.
        \textbf{(b)} \model demonstrates superior data efficiency, surpassing the fully-trained SFT baseline (100\% data) with only 25\% of the training data.
    }
    \label{fig:analysis_plots}
\end{figure}

As shown in Figure~\ref{fig:analysis_plots}(a), increasing the group size $G$ is critical for enhancing the reliability of policy advantage estimation in GRPO.
We observe a sharp performance boost as $G$ rises to 8, where the model effectively stabilizes its reasoning paths.
Beyond this point ($G>8$), performance gains exhibit diminishing returns, indicating that a moderate group size is sufficient to capture diverse reasoning patterns without excessive computational overhead.
Consequently, we select $G=8$ as the default setting to optimally balance accuracy with training throughput.

Figure~\ref{fig:analysis_plots}(b) illustrates a distinct contrast in learning dynamics.
While the SFT baseline exhibits gradual growth due to the difficulty of mapping raw inputs to constraints without massive data, \model demonstrates rapid adaptation.
Remarkably, \model trained on just 25\% of the data reaches 43.5\% Hit@1, surpassing the fully-trained SFT baseline by 12.26 percentage points (43.5\% vs. 31.24\%).
This result validates that our rubric-based reasoning acts as a strong structural prior, providing dense supervisory signals that enable robust generalization more efficiently than standard supervision.

\section{Additional Analyses and Discussions}
\label{appendix:additional_analyses}

\subsection{Supervision Quality}
\label{appendix:audits}

\paragraph{Rubric Facet Coverage.}
The four facets in Eq.~\ref{eq:rubric_set} follow SCR's core requirements: \textit{Situatedness} checks visual and spatial grounding, \textit{Item Relevance} checks user constraints and item attributes, \textit{Consistency} checks dialogue history and intent shifts, and \textit{Auxiliary} covers remaining domain-specific criteria. To verify their coverage, we manually checked 50 randomly sampled dialogue turns against the induced rubric set.

\begin{table}[ht]
\centering
\small
\begin{tabular}{lc}
\toprule
\textbf{Coverage} & \textbf{Dialogue Turns} \\
\midrule
Covered by the three core facets alone & 18/50 \\
Also required the Auxiliary facet & 32/50 \\
Contained an unmapped requirement & 0/50 \\
\bottomrule
\end{tabular}
\caption{Manual coverage check of the induced rubric facets on 50 sampled dialogue turns. All turn-level requirements map onto the four facets.}
\label{tab:facet_coverage}
\end{table}

\paragraph{Selector and Judge Decision Quality.}
To audit whether the training-time selector and reward judge make reliable decisions, we use \texttt{gpt-5.6-sol} to evaluate 200 sampled training turns, while three human annotators evaluate a 30-turn subset using the same 0-to-10 criteria. For both selector and judge decisions, 0 denotes an incorrect or unsupported decision, 5 a partially correct decision with missed or misapplied criteria, and 10 a fully correct decision supported by the scene and dialogue evidence. Scores are averaged across sampled turns and, for the human audit, across the three annotators.

\begin{table}[ht]
\centering
\small
\setlength{\tabcolsep}{3pt}
\begin{tabular}{lccc}
\toprule
\textbf{Evaluator} & \makecell{\textbf{Sample} \\ \textbf{Size}} & \makecell{\textbf{Selector} \\ \textbf{Quality} $\uparrow$} & \makecell{\textbf{Judge} \\ \textbf{Quality} $\uparrow$} \\
\midrule
\texttt{gpt-5.6-sol} & 200 & 9.2 & 8.8 \\
Three human annotators & 30 & 8.8 & 7.5 \\
\bottomrule
\end{tabular}
\caption{Decision-quality audit of the training-time rubric selector and reward judge on a 0-to-10 scale.}
\label{tab:selector_judge_audit}
\end{table}

Both the model and human audits indicate generally high decision quality for the selector and the reward judge (Table~\ref{tab:selector_judge_audit}). Since all reported end-to-end results use the model's own predicted preference states and resolved items without manual correction, residual errors of these components are already reflected in the reported results.

\paragraph{Negative-Sample Quality.}
The negative-synthesis prompts (Appendix~\ref{app:neg_prompts}) require introducing the intended conflict without changing unrelated content; satisfying both requirements constitutes a joint pass. With no separate filtering stage during construction, we audit quality post hoc using \texttt{gpt-5.6-sol} on 100 pairs per negative type and three human annotators on 15 pairs per type. The human annotators additionally vote on whether the gold response $Y_w$ is clearly preferable to the synthesized negative.

\begin{table}[ht]
\centering
\small
\setlength{\tabcolsep}{2pt}
\resizebox{\linewidth}{!}{
\begin{tabular}{lcc}
\toprule
\textbf{Evaluation} & \makecell{\textbf{Situation-} \\ \textbf{Conflicting}} & \makecell{\textbf{Preference-} \\ \textbf{Conflicting}} \\
\midrule
\texttt{gpt-5.6-sol} joint-pass (100/type) & 93.0\% & 88.0\% \\
Human joint-pass (15/type) & 86.7\% & 93.3\% \\
Human majority: $Y_w$ clearly preferred & 93.3\% & 86.7\% \\
\bottomrule
\end{tabular}}
\caption{Post-hoc quality audit of the synthesized heterogeneous negative samples.}
\label{tab:negative_audit}
\end{table}

Both audits show high pass rates for the intended conflicts, and a majority of the human annotators preferred the gold response for both negative types (Table~\ref{tab:negative_audit}).

\subsection{Robustness of Judge and User Simulator}
\label{appendix:evaluator_robustness}

\paragraph{Independent Response Judges.}
Since GPT-5.2 is used both for offline data construction and as the response evaluator, its judge scores may carry self-preference bias. We therefore rescore the same anonymized SIMMC 2.1 outputs with Gemini 3.1 Pro and Claude Opus 4.8, neither of which is used to train the evaluated \model configuration. Confidence intervals use a paired dialogue-cluster bootstrap.

\begin{table}[ht]
\centering
\small
\begin{tabular}{lc}
\toprule
\textbf{Response Judge} & \makecell{$\Delta$\textbf{Judge-Avg} (\model $-$ ReGeS) \\ \textbf{95\% CI}} \\
\midrule
GPT-5.2 (original) & +0.82 [0.65, 0.99] \\
Gemini 3.1 Pro & +0.59 [0.44, 0.76] \\
Claude Opus 4.8 & +0.90 [0.85, 0.97] \\
\bottomrule
\end{tabular}
\caption{Rescoring the same anonymized SIMMC 2.1 outputs with independent response judges. All three 95\% confidence intervals remain above zero.}
\label{tab:independent_judges}
\end{table}

As shown in Table~\ref{tab:independent_judges}, all three 95\% confidence intervals remain entirely above zero, so each evaluator assigns \model a positive Judge-Avg margin over ReGeS.

\paragraph{Alternative User Simulator.}
We further replace the original GPT-5.2 user simulator with Claude Opus 4.8 and evaluate the same 100 initial SIMMC 2.1 samples, with ten rollouts per sample and 1{,}000 simulated sessions per system, under the same interaction and termination rules described in Appendix~\ref{appendix:user_simulator}.

\begin{table}[ht]
\centering
\small
\setlength{\tabcolsep}{4pt}
\begin{tabular}{lcccc}
\toprule
\multirow{2}{*}{\textbf{User Simulator}} & \multicolumn{2}{c}{\textbf{ReGeS}} & \multicolumn{2}{c}{\textbf{\model}} \\
\cmidrule(lr){2-3} \cmidrule(lr){4-5}
 & \textbf{SR} $\uparrow$ & \textbf{VHR} $\downarrow$ & \textbf{SR} $\uparrow$ & \textbf{VHR} $\downarrow$ \\
\midrule
GPT-5.2 & 38.6 & 15.2 & \textbf{46.8} & \textbf{5.2} \\
Claude Opus 4.8 & 41.2 & 13.5 & \textbf{47.8} & \textbf{5.2} \\
\bottomrule
\end{tabular}
\caption{Dialogue-level evaluation on SIMMC 2.1 under two different user simulators. \model remains better than ReGeS under both simulators.}
\label{tab:simulator_replacement}
\end{table}

\paragraph{Direct Comparison with the Teacher-Scale Model.}
To quantify the gap between our 8B system and a much larger model, we evaluate GPT-5.2 directly with the same inputs and output format as the other inference-only methods on SIMMC 2.1. The few-shot setting uses the same three fixed training examples for every test instance; Hit@1 uses the predicted item, and Claude Opus 4.8 evaluates response quality to avoid self-scoring.

\begin{table}[ht]
\centering
\small
\begin{tabular}{lcc}
\toprule
\textbf{Method on SIMMC 2.1} & \textbf{Hit@1} $\uparrow$ & \makecell{\textbf{Opus 4.8} \\ \textbf{Judge-Avg}} $\uparrow$ \\
\midrule
GPT-5.2 zero-shot & 44.37 & 8.56 \\
GPT-5.2 few-shot & \textbf{51.85} & \textbf{8.83} \\
\model (Qwen3-VL-8B) & 48.12 & 8.73 \\
\bottomrule
\end{tabular}
\caption{Direct comparison with prompted GPT-5.2 on SIMMC 2.1. \model falls between zero-shot and few-shot GPT-5.2 on both measures.}
\label{tab:gpt_reference}
\end{table}

As shown in Table~\ref{tab:gpt_reference}, \model falls between zero-shot and few-shot GPT-5.2 on both measures despite using an 8B backbone. We treat prompted GPT-5.2 as a large-model reference rather than a formal upper bound, since prompting strategies and decoding configurations were not exhaustively tuned.

\subsection{Matched-Supervision Diagnostic}
\label{appendix:matched_supervision}

The main comparison in Table~\ref{table:recommendation_result} evaluates complete systems under their intended training objectives, and therefore does not by itself separate the contribution of our training objectives from the synthesized supervision used by \model. Adding rubric supervision or heterogeneous negative training directly to CRAG or ReGeS would create hybrid systems rather than preserve the published baselines.
To test whether the structured interface itself helps without any additional training data, we instead use each trained baseline's own frozen checkpoint to add a prompted first stage. This stage produces either a free-form rationale or a structured preference state, which is then provided to the same checkpoint for item recommendation and response generation. The checkpoint, call count, decoding configuration, and evaluation protocol are matched; only the intermediate representation differs.

\begin{table}[ht]
\centering
\small
\begin{tabular}{lcc}
\toprule
\textbf{Method on SIMMC 2.1} & \textbf{Hit@1} $\uparrow$ & \textbf{VHR} $\downarrow$ \\
\midrule
CRAG & 35.64 & 19.7 \\
CRAG + free-form reasoning & 37.02 & 18.1 \\
CRAG + preference state & 38.21 & 16.8 \\
\midrule
ReGeS & 39.45 & 15.2 \\
ReGeS + free-form reasoning & 40.71 & 14.1 \\
ReGeS + preference state & 42.14 & 12.7 \\
\midrule
\model & \textbf{48.12} & \textbf{5.2} \\
\bottomrule
\end{tabular}
\caption{Call-matched diagnostic on SIMMC 2.1 with the Qwen3-VL backbone. Each frozen baseline first generates an intermediate representation and then consumes it for recommendation and response generation.}
\label{tab:matched_supervision}
\end{table}

As shown in Table~\ref{tab:matched_supervision}, both baselines improve with free-form reasoning and improve further with the structured preference state under the same call budget, indicating that the structured interface contributes beyond synthesized training data.
Within \model, Fine-tuned CoT reaches 40.5\% Hit@1 versus 48.1\% for the full framework, while homogeneous DPO gives 13.4\% VHR and 11.2\% WMR versus 5.2\% and 7.4\% with heterogeneous negatives (\S\ref{sec:ablation}). These ablations show that rubric-guided reasoning and augmented alignment supervision also matter, but they do not fully separate optimization from data content or volume; we therefore do not attribute the full gain to GRPO or DPO alone.

\subsection{Preference Reasoner Distillation}
\label{appendix:distillation}

To reduce the inference latency introduced by explicit preference reasoning, we distill only the 8B preference reasoner into \texttt{Qwen/Qwen3-VL-4B-Instruct}. We use the same SIMMC 2.1 training contexts as for the 8B reasoner: the 8B model supplies structured preference-state targets, and the 4B student is trained with the same reasoning inputs, structural instruction, and output schema. At evaluation, only the reasoner checkpoint is replaced, while all other components remain unchanged; latency is measured with the same hardware and timing protocol as in Table~\ref{tab:latency_analysis}.
The distilled reasoner reduces total per-turn latency from 2342.7 ms to 1838.4 ms while retaining most of the recommendation gain (46.71\% vs. 48.12\% Hit@1); its VHR increases from 5.2\% to 8.2\% but remains substantially below ReGeS at 15.2\%. This provides a lower-latency alternative to the 8B configuration, although the quality-latency tradeoff remains.




\subsection{Robustness to Teacher and Judge Replacement}
\label{appendix:model_replacement}

To test whether our conclusions depend on specific external models, we conduct two model-replacement experiments on SIMMC 2.1.
First, Claude Opus 4.8 replaces GPT-5.2 as the teacher for rubric induction and negative-pair construction, while Qwen3-VL-8B-Instruct remains the training-time selector and reward judge.
Second, InternVL3-8B replaces Qwen3-VL-8B-Instruct as the selector and reward judge, while GPT-5.2 remains the teacher.
The source examples, number of negative pairs, prompt content (identical to the templates in Appendices~\ref{appendix:constitutional_meta_instruction}--\ref{app:neg_prompts}), training schedule, and evaluation protocol remain fixed.

\begin{table*}[ht]
\centering
\small
\begin{tabular}{llcccc}
\toprule
\textbf{Rubric and Negative-Pair Teacher} & \textbf{Training-Time Selector and Judge} & \textbf{Hit@1} $\uparrow$ & \textbf{Recall@5} $\uparrow$ & \textbf{SR} $\uparrow$ & \textbf{VHR} $\downarrow$ \\
\midrule
GPT-5.2 (default) & Qwen3-VL-8B-Instruct (default) & 48.12 & 65.08 & 46.8 & 5.2 \\
Claude Opus 4.8 & Qwen3-VL-8B-Instruct & 49.54 & 68.60 & 48.2 & 4.9 \\
GPT-5.2 & InternVL3-8B & 44.46 & 61.42 & 42.6 & 9.8 \\
\bottomrule
\end{tabular}
\caption{Model-replacement experiments on SIMMC 2.1 with the Qwen3-VL backbone. Replacing the teacher or the selector/judge preserves the conclusions of the main experiments.}
\label{tab:model_replacement}
\end{table*}

As reported in Table~\ref{tab:model_replacement}, the Claude Opus 4.8 teacher improves all four metrics over the GPT-5.2 teacher, and the InternVL3-8B selector/judge lowers absolute performance but still clearly outperforms the strongest baseline ReGeS (39.45\% Hit@1, 15.2\% VHR). These results indicate that \model remains effective when a different teacher or judge model is used.

\subsection{Cross-Domain Transfer}
\label{appendix:cross_domain}

To examine the generalizability of the learned rubrics and models, we evaluate bidirectional zero-shot transfer between the Fashion and Furniture domains in SIMMC 2.1. In each direction, the complete system is trained on the source domain and evaluated on the target domain without any target-domain training or development data.

\begin{table}[t!]
\centering
\small
\begin{tabular}{llcc}
\toprule
\textbf{Train $\rightarrow$ Test} & \textbf{Method} & \textbf{Hit@1} $\uparrow$ & \textbf{VHR} $\downarrow$ \\
\midrule
\multirow{2}{*}{Furniture $\rightarrow$ Fashion} & ReGeS & 25.84 & 22.9 \\
 & \model & \textbf{32.17} & \textbf{10.8} \\
\midrule
\multirow{2}{*}{Fashion $\rightarrow$ Furniture} & ReGeS & 27.36 & 21.7 \\
 & \model & \textbf{33.94} & \textbf{10.2} \\
\bottomrule
\end{tabular}
\caption{Bidirectional zero-shot cross-domain transfer between the Fashion and Furniture domains of SIMMC 2.1. The relative ordering of \model and ReGeS is consistent in both directions.}
\label{tab:cross_domain}
\end{table}

As shown in Table~\ref{tab:cross_domain}, the relative ordering of \model and ReGeS is consistent in both directions despite the absence of target-domain data.
To distinguish which components transfer, we further examine the induced rubric criteria and the response generator separately in a distinct setting that permits 10\% target-domain adaptation data.
Reusing source-induced rubric criteria is within 0.57 and 0.65 Hit@1 percentage points of inducing new target-domain criteria on Fashion and Furniture, respectively.
Under identical target-domain contexts, the source-trained generators obtain Gemini 3.1 Pro Judge-Avg scores of 8.12 on Fashion and 8.34 on Furniture, compared with 9.16 and 9.20 for the corresponding target-trained generators.
These patterns suggest that the core rubric facets capture SCR requirements that transfer across product domains, while the Auxiliary facet accommodates domain-specific constraints; response generation is more domain-sensitive because item attributes and their linguistic realization change across domains.
The current evidence is limited to the Fashion and Furniture domains; broader cross-domain transfer remains future work.

\section{Case Study}
\label{app:case_study}

\begin{figure*}[t!]
    \centering
    \begin{tcolorbox}[
        enhanced,
        title={\small \bfseries \faImage \ \ Situated Conversation for Recommendation},
        colback=c_context_bg,
        colframe=c_context_frame,
        coltitle=white,
        boxrule=0.8pt,
        arc=3mm,
        drop shadow,
        sidebyside,
        sidebyside align=top,
        lefthand width=0.55\linewidth,
        sidebyside gap=5mm
    ]
        \centering
        \includegraphics[width=1.0\linewidth]{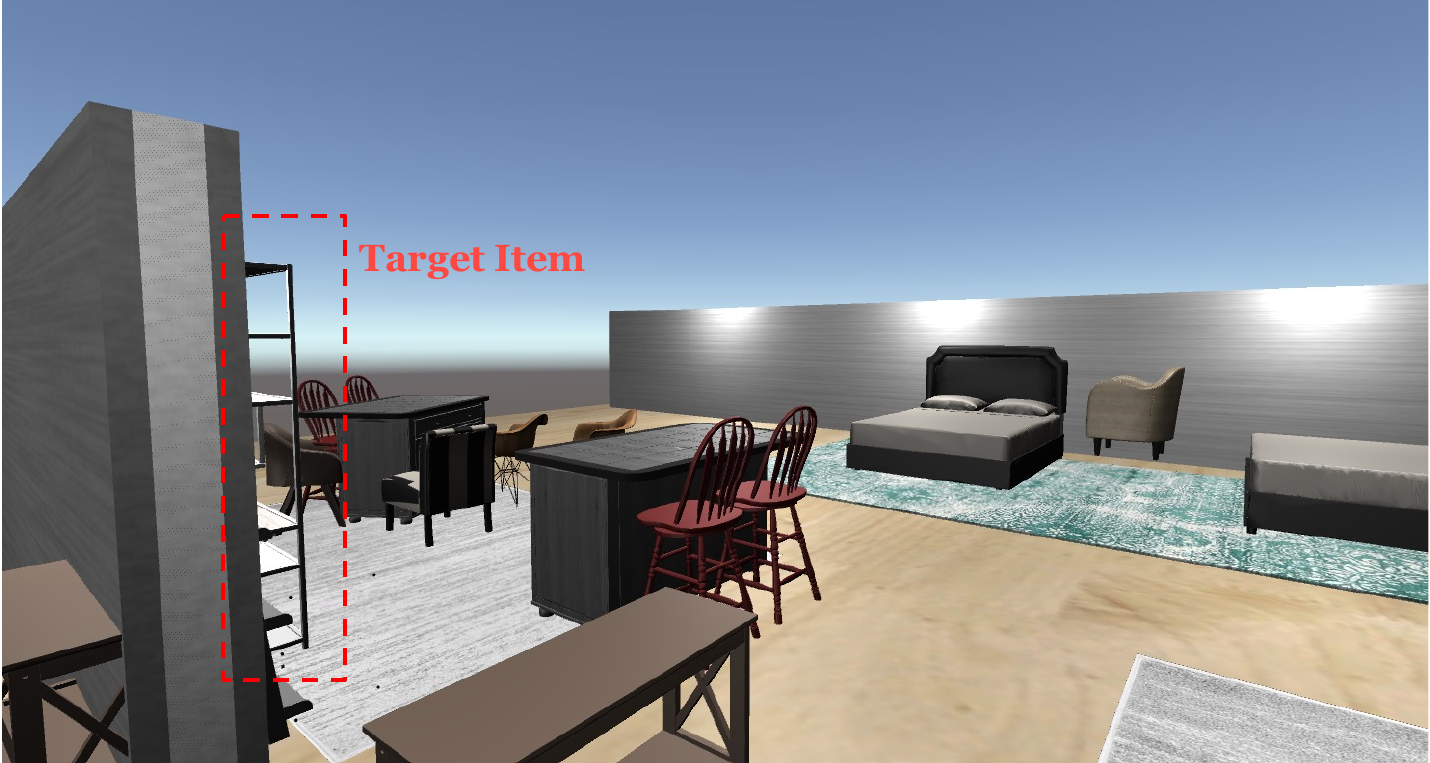}
        \vspace{2pt}
        \captionof*{figure}{\footnotesize \textbf{Scene:} Furniture store with a desk area (left) and bedroom (right).}
        
        \tcblower
        
        \small
        \renewcommand{\arraystretch}{1.3} 
        \begin{tabularx}{\linewidth}{@{} l >{\raggedright\arraybackslash}X @{}}
            \textbf{User:}  & Can you help me get a leather chair? \\
            \textbf{Agent:} & I have two leather chairs: a brown one and a black-and-white one by the kitchen island. \\
            \textbf{User:}  & \textbf{I also need extra storage space} to fit the ambiance of my new decor style. \\
            \textbf{Agent:} & I see several shelving units. Do you have any color or placement in mind? \\
            \textbf{User:}  & \textcolor{c_fail_frame}{\textbf{I don't want anything in that light wood finish.}} Is there a darker one near the desk?
        \end{tabularx}
    \end{tcolorbox}

    \vspace{-2pt}

    \begin{tcolorbox}[
        enhanced,
        title={\small \bfseries \faTimesCircle \ \ Baselines},
        colback=c_fail_bg,
        colframe=c_fail_frame,
        coltitle=white,
        boxrule=0.8pt,
        arc=3mm,
        drop shadow
    ]
        \footnotesize
        \textbf{\textit{1. Vanilla Prompting}} \\
        \textcolor{gray!40!black}{``The black-and-white leather chair is by the kitchen island...''} \\
        $\hookrightarrow$ \textit{Fails to recognize the task shift from chairs to storage units.}
        
        \vspace{4pt} \hrule \vspace{4pt}

        \textbf{\textit{2. ICL}} \\
        \textcolor{gray!40!black}{``I have a leather chair and a \textbf{white storage bin} for you...''} \\
        $\hookrightarrow$ \textit{Hallucinates a non-existent item (white bin) while ignoring the spatial constraint.}
        
        \vspace{4pt} \hrule \vspace{4pt}
        
        \textbf{\textit{3. CoT}} \\
        \textcolor{gray!40!black}{``I see a light wood shelf near the desk. It should fit your decor style.''} \\
        $\hookrightarrow$ \textit{Correctly grounds the location but ignores the explicit negation (``don't want light wood'').}
        
        \vspace{4pt} \hrule \vspace{4pt}

        \textbf{\textit{4. SFT}} \\
        \textcolor{gray!40!black}{``Sure! Here is a \textbf{light wood shelving unit} at the desk. It fits your decor style!''} \\
        $\hookrightarrow$ \textit{Exhibits sycophancy: blindly agrees with the user ('fits your style') while recommending an item that explicitly violates the negative constraint.}
        
        \vspace{4pt} \hrule \vspace{4pt}
        
        \textbf{\textit{5. ReGeS}} \\
        \textcolor{c_fail_frame}{\textbf{Pred:}} Light Wood Display Cabinet \\
        \textcolor{gray!40!black}{``How about this light wood cabinet? It offers extra storage and is near the desk.''} \\
        $\hookrightarrow$ \textit{Overfits to the visually salient light-colored furniture, failing to align with the negative preference.}
    \end{tcolorbox}

    \vspace{-2pt}

    \begin{tcolorbox}[
        enhanced,
        title={\small \bfseries \faCheckCircle \ \ \model (Ours)},
        colback=c_ours_bg,
        colframe=c_ours_frame,
        coltitle=white,
        boxrule=0.8pt,
        arc=3mm,
        drop shadow
    ]
        \footnotesize
        \textbf{\faBrain \ \ Preference Thought ($\mathcal{T}$):} \\
        \textit{``The user explicitly \textbf{rejects} the `light wood finish', implying a need for a darker material. The item must be a `storage unit' located in the `desk area' (left side)...''}
        
        \vspace{4pt} \hrule \vspace{4pt}
        
        \textbf{\faDatabase \ \ Preference State ($\mathcal{P}$):} 
        \begin{itemize}[leftmargin=1.2em, nosep, label={\tiny\faCaretRight}]
            \item \textbf{Intent:} Get Storage Unit
            \item \textbf{Constraint:} \textcolor{c_fail_frame}{$\neg$ Light Wood} $\land$ Near Desk
            \item \textbf{Visual Target:} \textcolor{c_ours_frame}{\textbf{Black Metal Shelving}}
        \end{itemize}
        
        \vspace{4pt} \hrule \vspace{4pt}
        
        \textbf{\faCommentDots \ Generated Response:} \\
        ``I found a \textbf{black metal shelving unit} next to the desk. It matches the darker ambiance and avoids the light wood finish...''
        
    \end{tcolorbox}

    \vspace{-5pt}
    \caption{Qualitative case study on a challenging scenario. While baselines struggle with visual saliency (ReGeS) or fail to reason about negation (CoT), our \model leverages structured preference reasoning to identify the target and generate a more contextually-appropriate response.}
    \label{fig:visual_case_study}
\end{figure*}

To illustrate the behavior of \model, we present a detailed qualitative comparison in Figure~\ref{fig:visual_case_study}. We select a challenging scenario from the SCREEN dataset that requires both \textbf{spatial grounding} (locating the desk area) and \textbf{negative constraint reasoning} (excluding specific visual attributes).

